\documentclass[letterpaper]{article} 
\usepackage{aaai2027} 
\nocopyright
\usepackage[hyphens]{url} 
\usepackage{graphicx} 
\usepackage{natbib} 
\usepackage{caption} 
\usepackage{booktabs}
\usepackage{amsmath}
\usepackage{amssymb}
\usepackage{xpatch}
\newcommand{\method}{Tactile-WAM}
\newcommand{\videoclean}{\textsc{VideoClean}}
\newcommand{\taam}{\textsc{TAAM}}
\newcommand{\cmark}{\ensuremath{\checkmark}}

\makeatletter
\newcommand{\firstpageoverview}{%
  \par\vspace{-0.45in}%
  \begin{minipage}{\textwidth}
    \centering
    \includegraphics[width=0.85\textwidth]{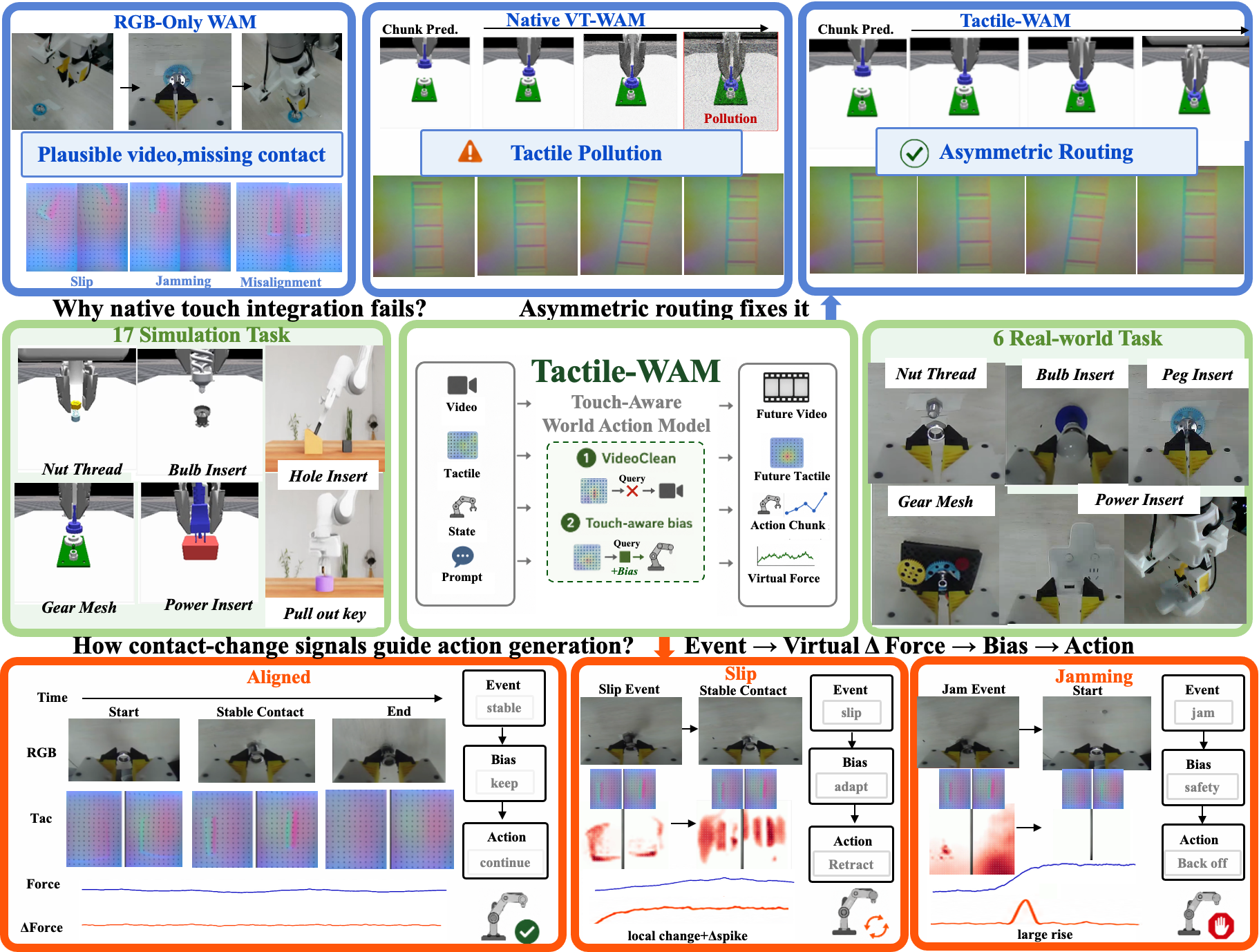}
    \captionof{figure}{System overview. The top row contrasts RGB-only prediction, naive visuo-tactile modeling, and asymmetric routing; the middle summarizes the model inputs, outputs, and task suites; the bottom shows aligned, slip, and jamming contact traces.}
    \label{fig:motivation}
  \end{minipage}%
  \vspace{0.02in}%
}
\xpatchcmd{\maketitle}
  {\twocolumn[\@maketitle]}
  {\twocolumn[\@maketitle\firstpageoverview]}
  {}
  {\PackageError{tactile-wam}{Unable to place the first-page overview}{}}
\makeatother

\title{\method: Touch-Aware World Action Model with Tactile Asymmetric Attention}
\author{
Siyu Wu\textsuperscript{1,4,*},
Linjing You\textsuperscript{1,2,3,*},
Junjie Zhu\textsuperscript{1,\dag},
Yaozu Liu\textsuperscript{2},
Huang Kaixiang\textsuperscript{4},
Chen Yonghang\textsuperscript{4},
Jituo Li\textsuperscript{4},
Changhao Zhang\textsuperscript{1},
Jian Liu\textsuperscript{1},
Hengshuo Chu\textsuperscript{1},
Qi Li\textsuperscript{2},
Hengshuang Zhao\textsuperscript{3}
}

\affiliations{
\textsuperscript{1}Ant Group\\
\textsuperscript{2}Institute of Automation, Chinese Academy of Sciences\\
\textsuperscript{3}The University of Hong Kong\\
\textsuperscript{4}Zhejiang University\\
\textsuperscript{*}Equal contribution\\
\textsuperscript{\dag}Corresponding author: yuxue.zjj@antgroup.com
}

\begin{document}

\maketitle

\begin{abstract}
World Action Models (WAMs) jointly predict future visual observations and actions, but visual futures alone often miss slip, jamming, contact-direction changes, and subtle misalignment in contact-rich manipulation. Tactile signals reveal these hidden physical states, yet naive tactile-token injection can disrupt visual dynamics modeling due to the limited scale of tactile data, a phenomenon we term \textbf{tactile pollution}. We introduce \textbf{Tactile-WAM}, which uses asymmetric attention to block video queries from tactile keys while preserving tactile access for action queries. A contact-change-aware bias further strengthens action attention to touch. Because tactile pixel changes do not reliably reflect contact changes, we derive a six-dimensional touch-aware proxy from tactile optical flow. Observed proxy changes drive the attention bias, while future-proxy supervision preserves action-relevant contact dynamics in predicted tactile representations. On ManiFeel, visual-path isolation reduces deviation from the RGB-only trajectory by 21.8\% in MSE at the step-matched 20K checkpoint without a statistically detectable change in ground-truth video quality. The full model improves average success from 15.6\% to 32.7\%, with \videoclean{} providing the largest gain. On five real-robot tasks, \method{} achieves 49.2\% success.
\end{abstract}

\section{Introduction}

World Action Models (WAMs) jointly predict future states and robot actions, enabling policies to anticipate the consequences of an action sequence~\citep{dreamzero2026,pad2024,vpp2024}. Through large-scale video pretraining, they acquire strong visual appearance and motion priors. However, their predicted futures remain predominantly visual and cannot fully capture the physical states involved in contact-rich manipulation.

In tasks such as insertion, assembly, and object reorientation, success often depends on slip, jamming, contact-direction changes, and subtle misalignment. These events may be difficult to observe from RGB images but directly determine the next corrective action, as illustrated in Figure~\ref{fig:motivation}. Vision-based tactile sensors capture local deformation and contact transitions~\citep{gelsight2017,digit2020,tactip2018}, and learned tactile representations have shown clear benefits in contact-rich manipulation~\citep{unit2024,sparsh2024,rdp2025,manifeel2025}. A WAM for such tasks should therefore predict future tactile states and effectively use touch for action generation.

Directly incorporating touch, however, is not always beneficial. Tactile datasets are far smaller than video pretraining corpora, and tactile signals mainly describe sparse, local contact events. Unconstrained attention from video queries to tactile keys can therefore interfere with the pretrained visual representation and degrade both video and action prediction. We term this phenomenon \emph{tactile pollution}. As shown in Figure~\ref{fig:motivation}, naive tactile fusion produces blur and distortion in predicted visual futures.

We further observe that pixel-level changes in tactile images do not reliably reflect actual contact changes. Small pixel differences may correspond to critical events such as slip, compression, or slight in-gripper rotation, while large pixel differences may result from visually prominent but physically minor variations~\citep{piphen2026,manilongshot2026}. Our analysis shows only weak correlation between tactile pixel similarity and deformation-derived contact similarity. Thus, effective tactile modeling must determine when touch is important and preserve contact-relevant dynamics in predicted tactile representations.

Motivated by these observations, we introduce \method{}, a Wan2.2-based tactile-aware world action model. Its core component, the Tactile Asymmetric Attention Mechanism (\taam{}), prevents video queries from attending to tactile keys while preserving tactile access for action queries. We further derive a six-dimensional touch-aware proxy from optical flow between consecutive tactile images. Observed proxy changes generate a causal attention bias that strengthens action attention to touch, while future-proxy supervision encourages predicted tactile representations to preserve action-relevant contact dynamics.

Our main contributions are as follows:
\begin{itemize}
\item We identify and quantify \emph{tactile pollution} in visually pretrained WAMs and introduce \taam{} to protect video prediction while retaining tactile information for action generation;
\item We reveal the mismatch between tactile pixel changes and contact changes, and propose an optical-flow-based touch-aware proxy for adaptive attention and future tactile supervision;
\item We validate \method{} on nine simulation tasks and five real-robot tasks, demonstrating substantial improvements in contact-rich manipulation while preserving visual prediction quality.
\end{itemize}

\section{Preliminaries and Notation}
\label{sec:preliminaries}

We briefly review World Action Models (WAMs) and vision-based tactile sensing to establish the notation used throughout this paper. Detailed related work is deferred to Appendix~A.

\subsection{World Action Models}

WAMs jointly predict future visual observations and robot actions, conditioning action generation on environment evolution~\citep{dreamzero2026,pad2024,vpp2024}. At policy-call time $t$, a WAM receives RGB history $o^v_{\leq t}$, proprioceptive states $s_{\leq t}$, and language instruction $\ell$, collectively denoted as
\begin{equation}
\mathcal{C}_t = \left(o^v_{\leq t}, s_{\leq t}, \ell\right).
\label{eq:wam_context}
\end{equation}

Let $E_v$ encode a future RGB sequence $o^v_{t+1:t+H}$ (horizon $H$) into latent $z_0^v = E_v(o^v_{t+1:t+H})$, and let action chunk $a_0 = a_{t:t+K-1}$ ($K$: action horizon).

Following conditional flow matching~\citep{flowmatching2023}, WAMs transform Gaussian noise into future latents and actions. For each target $x_0^m$ ($m\in\{v,a\}$), with noise $\epsilon^m\sim\mathcal{N}(0,I)$ and timestep $\rho\in[0,1]$, the interpolant and velocity are
\begin{equation}
x_\rho^m = (1-\rho)x_0^m + \rho\epsilon^m,\qquad
u_\rho^m = \epsilon^m - x_0^m,
\label{eq:flow_path}
\end{equation}
where $x_0^v=z_0^v$ and $x_0^a=a_0$. Conditioned on $\mathcal{C}_t$, the model predicts vector field $v_\theta^m$ via
\begin{equation}
\mathcal{L}_{\mathrm{FM}}^m =
\mathbb{E}\!\left[
\left\|
v_\theta^m(x_\rho^m,\rho;\mathcal{C}_t) - u_\rho^m
\right\|_2^2
\right].
\label{eq:flow_matching}
\end{equation}
Inference transforms sampled noise into future visual latents and executable actions.

\subsection{Vision-Based Tactile Sensing}

Vision-based tactile sensors capture contact-induced surface deformation via an internal camera~\citep{gelsight2017,digit2020,tactip2018}. Let
\begin{equation}
o_t^{\tau,r} \in \mathbb{R}^{H_\tau \times W_\tau \times 3},\qquad r\in\{L,R\},
\label{eq:tactile_image}
\end{equation}
be the tactile image from the left/right sensor at time $t$, with history $o_{\leq t}^{\tau,r}$. These images reflect local contact geometry, pressure, and temporal dynamics (e.g., shear, slip), but—unlike calibrated force measurements—are high-dimensional visual observations that do not directly yield physical force values.

\section{Key Observations}

\subsection{Tactile Pollution}

\begin{figure}[!t]
    \centering
    \includegraphics[width=\linewidth,height=0.842\linewidth,keepaspectratio]{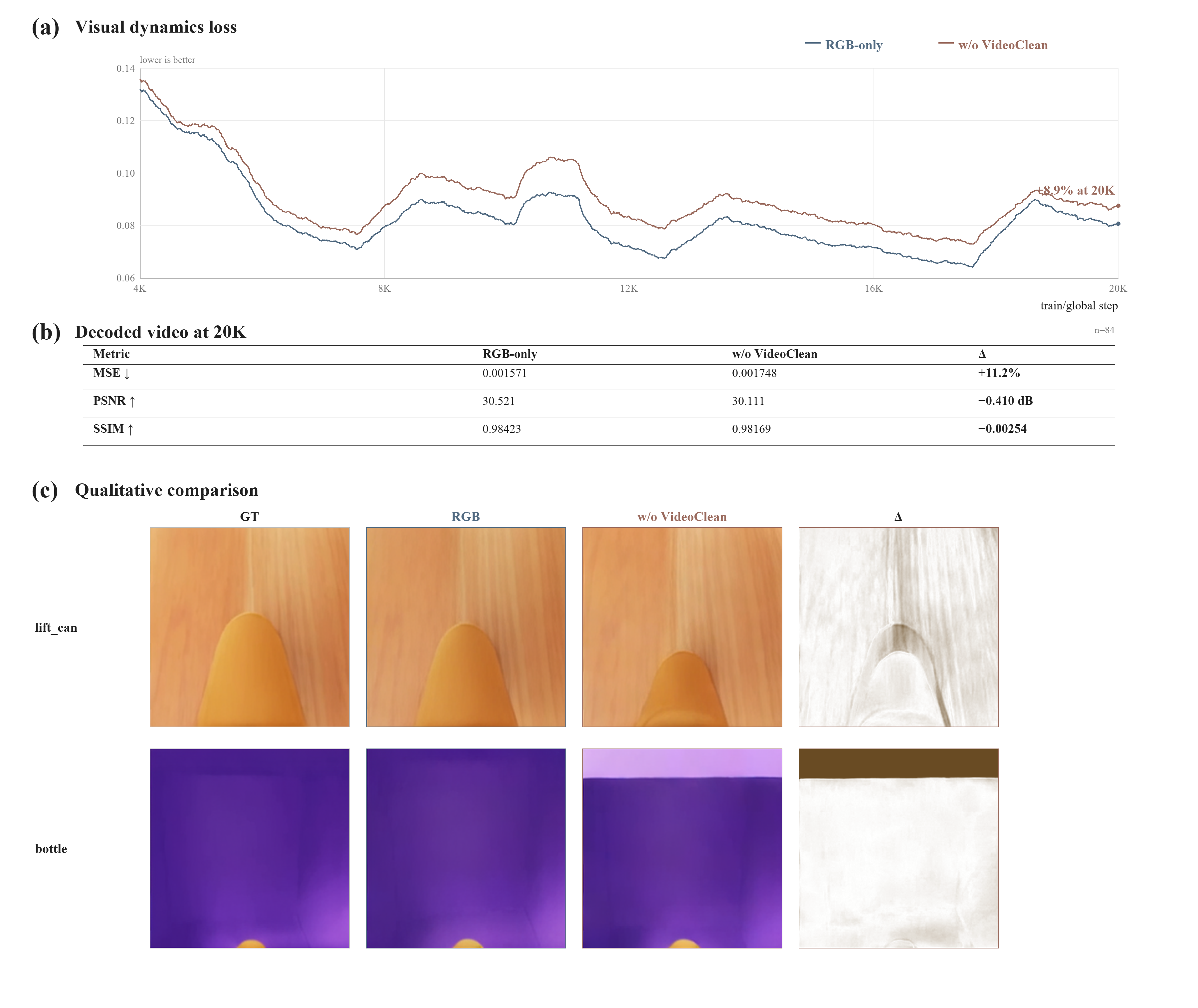}
    \caption{Tactile pollution on UniVTAC. (a) Training loss. (b) Prediction MSE vs.~ground truth. (c) Qualitative comparison.}
    \label{fig:sim_results}
\end{figure}

A straightforward approach to incorporate tactile information into WAMs is to jointly model tactile observations with visual inputs and actions. In DiT-based architectures, tokens from all modalities are concatenated into a shared sequence, allowing visual queries to attend to tactile keys. For visual tokens, the attention operation is:
\begin{equation}
\mathrm{Attn}(Q^v,K,V)
=
\mathrm{Softmax}
\left(
\frac{Q^vK^\top}{\sqrt{d}}
\right)V,
\label{eq:naive_attention}
\end{equation}
where $Q^v$ accesses key-value representations from all modalities:
\begin{equation}
K,V=
\left[
K^v,K^\tau,K^a,K^s
\right],
\end{equation}
with $K^v,K^\tau,K^a,K^s$ denoting keys of visual, tactile, action, and state tokens, respectively. We refer to this direct fusion as \emph{Naive VT-WAM}.

However, unrestricted tactile injection degrades visual prediction and action generation. On UniVTAC (800 training trajectories), Naive VT-WAM consistently exhibits higher training loss than DreamZero (Fig.~2(a)), yields ~50\% higher MSE in future frame prediction (Fig.~2(b)), and produces visible blur around manipulated objects (Fig.~2(c)). These results indicate that fine-tuning a visually pretrained WAM with limited tactile data interferes with the visual dynamics prior. We term this \emph{Tactile Pollution}.

\subsection{Pixel--Contact Misalignment}

\begin{figure}[t]
    \centering
    \includegraphics[width=\linewidth]{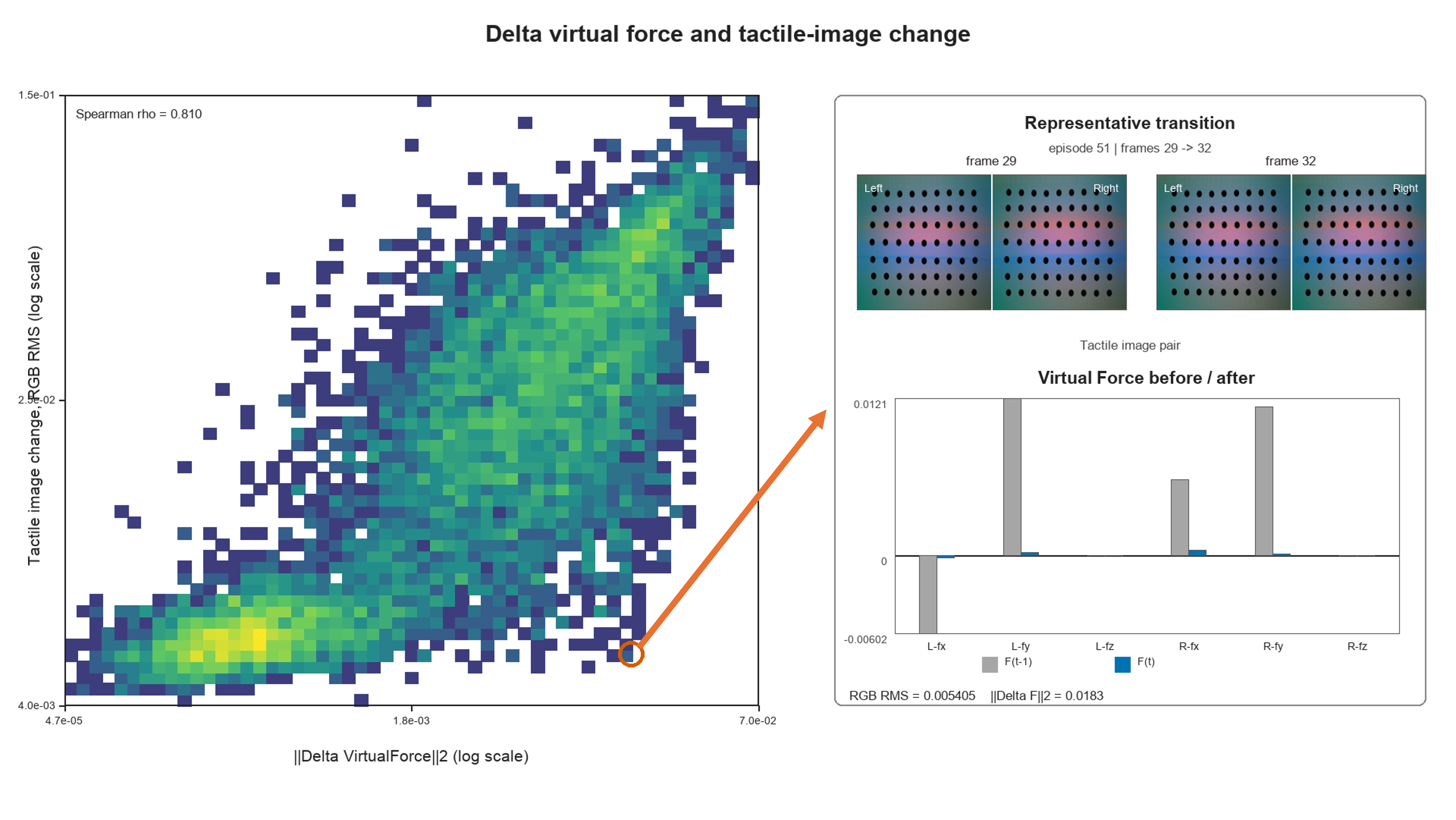}
    \caption{Pixel–contact misalignment. Each point denotes a pair of adjacent tactile images; x-axis: contact variation, y-axis: pixel variation.}
    \label{fig:latent_contact}
\end{figure}

We further question whether pixel reconstruction—effective for RGB video generation—is suitable for tactile prediction. Unlike visual appearance, tactile sensing primarily captures local contact states and their transitions (slip, compression, object motion), making pixel-level changes an unreliable proxy for contact dynamics.

To verify this, we sample 5,000 tactile image pairs and analyze pixel variation against contact variation (Fig.~\ref{fig:latent_contact}). No clear linear correlation is observed. Samples in the blue region show large pixel changes despite limited contact variation, while the red region exhibits the opposite. Samples on the yellow dashed line share similar pixel variations but differ substantially in contact changes, and those on the green dashed line show the reverse.

These observations indicate that tactile representation space does not reliably reflect contact dynamics. Thus, pixel reconstruction objectives alone cannot ensure predicted tactile representations preserve contact changes critical for action generation.

\section{Method}
\label{sec:method}

\subsection{Architecture Overview}
\label{subsec:overview}

\begin{figure*}[t]
    \centering
    \includegraphics[width=0.98\textwidth]{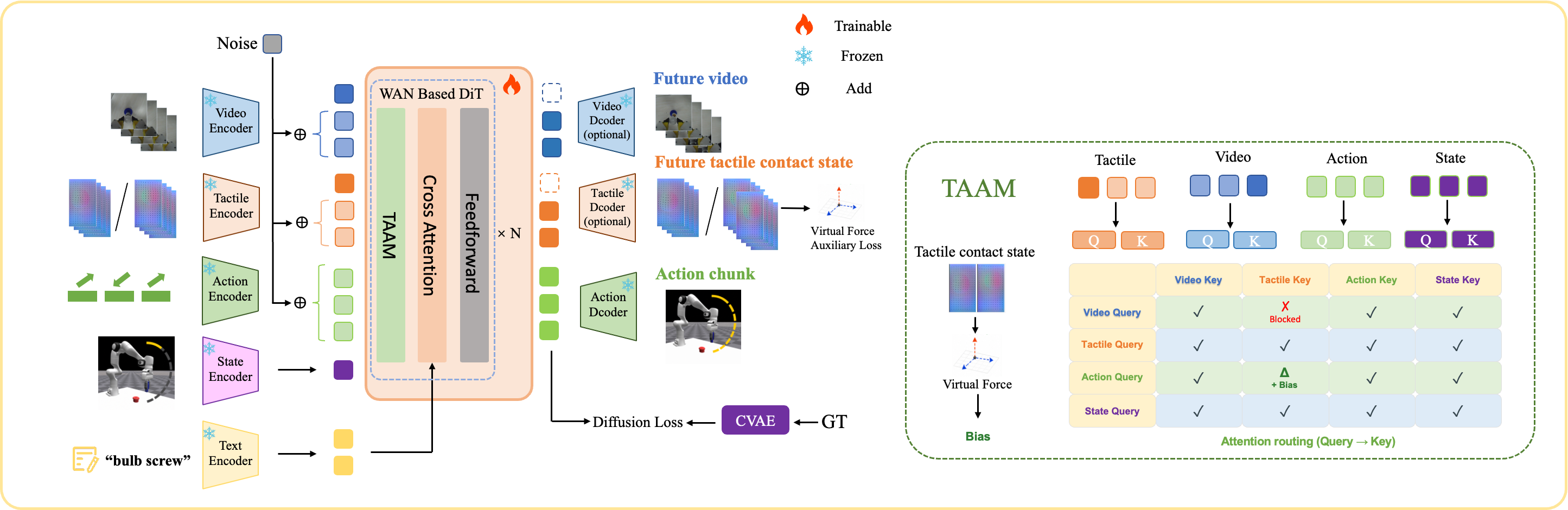}
    \caption{Overview of \textsc{Tactile-WAM}. RGB and tactile histories are encoded by a shared frozen VAE, and a Wan DiT jointly predicts future visual latents, tactile latents, and action chunks. A \textsc{Videoclean} mask prevents visual queries from accessing tactile keys, while a touch-aware bias routes action queries to tactile features within each causal block. A differentiable deformation proxy supervises tactile prediction and drives the bias computation.}
    \label{fig:method}
\end{figure*}

Figure~\ref{fig:method} illustrates the overall framework of \textsc{Tactile-WAM}. RGB and tactile history sequences are independently encoded into separate visual and tactile latent variables via the frozen Wan2.2 VAE, with corresponding modality embeddings added. The language-conditioned Wan DiT then jointly predicts future video frames, future tactile states, and action chunks within a unified framework.

The remainder of this section is organized as follows. We first describe the joint visual-tactile-action modeling based on Wan (Section 4.2). To address the tactile pollution and pixel–touch misalignment issues identified in Section 4.3, we introduce a tactile asymmetric attention mask (Section 4.4) and a touch-aware proxy (Section 4.5), respectively. Specifically, the mask module blocks visual queries from accessing tactile keys to mitigate tactile pollution, while simultaneously encouraging action queries to attend more to tactile information during contact changes via a touch-aware attention bias. To resolve pixel–touch misalignment, we employ a touch-aware proxy to derive the attention bias and additionally supervise the predicted tactile images with proxy signals during training. Further architectural details and training configurations are provided in Appendix~B.

\subsection{Joint Visual-Tactile-Action Modeling}
\label{subsec:joint_modeling}

To incorporate vision-based tactile sensing into the unified world-action-model framework, tactile observations are first encoded into a latent space shared with vision. Specifically, the left and right tactile images \(o^{\tau,L}_t, o^{\tau,R}_t\) (Eq.~(6)) are resized, horizontally mosaicked via \(\mathcal M\), and encoded by the causal VAE encoder \(E_{\mathrm{Wan}}\) of the Wan2.2 backbone:
\begin{equation}
z^\tau = E_{\mathrm{Wan}}\!\left(\mathcal M(o^{\tau,L},o^{\tau,R})\right),
\label{eq:tactile_enc}
\end{equation}
where the future prediction target is denoted as \(z_0^\tau\), defined in parallel with the visual latent \(z_0^v\) in Eq.~(2).

The conditional context is accordingly extended to explicitly incorporate the left and right tactile histories:
\begin{equation}
\mathcal{C}_t^\tau = \left(o^v_{\leq t},\ \{o^{\tau,r}_{\leq t}\}_{r=L,R},\ s_{\leq t},\ \ell\right),
\label{eq:context_tactile}
\end{equation}
where historical tactile latents, after learnable linear projection and modality embedding, are injected into the denoising network as auxiliary conditional signals.

Training and inference for the tactile modality directly follow the flow-matching paradigm defined in Section~\ref{sec:preliminaries}: substituting the conditional context in Eq.~(4) with \(\mathcal{C}_t^\tau\) yields the tactile flow-matching loss \(\mathcal{L}_{\mathrm{FM}}^\tau\); during inference, the clean latent \(\widehat z_0^\tau\) is reconstructed from the sampled noise \(\epsilon^\tau\) following Eq.~(5), and then decoded by \(D_{\mathrm{Wan}}\) to produce the predicted tactile images. Each action chunk is temporally aligned with the corresponding tactile latent frames. Finally, the flow-matching losses for vision, action, and tactile modalities are jointly optimized.

\subsection{Tactile Asymmetric Attention Mask}
\label{subsec:tactile_asymmetric_mask}

Naive symmetric routing allows visual queries to directly attend to sparse tactile keys. Since local tactile events do not necessarily predict global future appearance, this pathway can corrupt the pre-trained visual dynamics representation, leading to the tactile pollution failure mode. To address this, we propose a tactile asymmetric attention mask that protects visual prediction while retaining tactile information for action generation.

\paragraph{\textsc{Videoclean} Mask.}
Let \(G(q)\) and \(G(k)\) denote the token groups to which query \(q\) and key \(k\) belong. We block only the access from visual queries to tactile keys:
\begin{equation}
M^{\mathrm{vc}}_{qk}=
\begin{cases}
-\infty,&G(q)=V\ \wedge\ G(k)=T,\\
0,&\text{otherwise}.
\end{cases}
\label{eq:videoclean}
\end{equation}
Action queries retain full access to tactile keys, and tactile queries preserve their multimodal context. Thus, this mask protects the visual prediction pathway in a unidirectional manner, rather than completely isolating modalities.

\begin{table*}[h]
\centering
\scriptsize
\setlength{\tabcolsep}{3.2pt}
\renewcommand{\arraystretch}{1.08}
\caption{UniVTAC success counts after 100K training steps, with 20 trials per task. Bold marks the best result per task.}
\label{tab:univtac_results}
\resizebox{\textwidth}{!}{
\begin{tabular}{lccccccccc}
\toprule
Method & Grasp cls. & Hole ins. & Tube ins. & HDMI ins. & Lift bottle & Lift can & Pull key & Bottle shelf & Overall \\
\midrule
\method{} & $11/20$ & $4/20$ & $\mathbf{17/20}$ & $0/20$ & $1/20$ & $2/20$ & $4/20$ & $9/20$ & $48/160$ (30.0\%) \\
w/o \videoclean{} & $10/20$ & $0/20$ & $12/20$ & $\mathbf{4/20}$ & $6/20$ & $4/20$ & $4/20$ & $2/20$ & $42/160$ (26.3\%) \\
DreamZero & $16/20$ & $4/20$ & $4/20$ & $2/20$ & $2/20$ & $2/20$ & $2/20$ & $2/20$ & $34/160$ (21.3\%) \\
$\pi_{0.5}$ & $\mathbf{19/20}$ & $\mathbf{8/20}$ & $4/20$ & $0/20$ & $\mathbf{11/20}$ & $\mathbf{8/20}$ & $\mathbf{9/20}$ & $\mathbf{10/20}$ & $\mathbf{69/160}$ (43.1\%) \\
\bottomrule
\end{tabular}}
\end{table*}

\begin{table*}[h]
\centering
\scriptsize
\setlength{\tabcolsep}{3.1pt}
\renewcommand{\arraystretch}{1.08}
\caption{ManiFeel success counts after 60K training steps, with 50 trials per task. Bold marks the best result per task.}
\label{tab:manifeel_results}
\resizebox{\textwidth}{!}{
\begin{tabular}{lcccccccccc}
\toprule
Method & Peg ins. & USB ins. & Power ins. & Gear ins. & Bulb ins. & Bolt-nut & Object search & Peg reorient. & Ball sort & Overall \\
\midrule
\method{} & $\mathbf{8/50}$ & $1/50$ & $9/50$ & $\mathbf{19/50}$ & $\mathbf{34/50}$ & $\mathbf{20/50}$ & $2/50$ & $17/50$ & $37/50$ & $147/450$ (32.7\%) \\
DreamZero & $4/50$ & $0/50$ & $4/50$ & $9/50$ & $17/50$ & $10/50$ & $0/50$ & $8/50$ & $18/50$ & $70/450$ (15.6\%) \\
$\pi_{0.5}$ & $5/50$ & $\mathbf{5/50}$ & $\mathbf{14/50}$ & $11/50$ & $18/50$ & $15/50$ & $\mathbf{28/50}$ & $\mathbf{21/50}$ & $\mathbf{48/50}$ & $\mathbf{165/450}$ (36.7\%) \\
\bottomrule
\end{tabular}}
\end{table*}

\paragraph{Touch-aware Attention Bias.}
To enable action prediction to perceive the current contact state, we design a touch-aware attention bias that guides action queries to attend to tactile representations within the same temporal block. The tactile sequence is partitioned into causal blocks according to the action-chunk boundaries, where each block relies solely on historical observations up to its end time to ensure causality.

For each tactile causal block \(c\), we extract a touch-aware perception vector from the tactile frames at the previous and current policy-calling instants:
\begin{equation}
F^{\mathrm{obs}}_c=\Phi(o^\tau_{u_c-8},o^\tau_{u_c}),
\label{eq:observed_force}
\end{equation}
where \(\Phi(\cdot)\) denotes the differentiable deformation estimator detailed in Section~\ref{subsec:proxy}, and \(u_c\) is the anchor timestamp of the block. To obtain a compact gating signal, we convert this high-dimensional vector into a contact intensity score:
\begin{equation}
s_c=\tanh\!\left(\frac{\operatorname{ReLU}(\|F^{\mathrm{obs}}_c\|_2-\theta)}{T}\right),
\label{eq:force_score}
\end{equation}
which is normalized to \([0,1)\) via threshold \(\theta\) and temperature \(T\), reflecting the presence and strength of contact. Each tactile patch key \(k\) within the block receives the following bias enhancement:
\begin{equation}
b_{c,k}=\operatorname{clip}\!\left(\alpha s_c,0,b_{\max}\right),
\label{eq:bias_map}
\end{equation}
where \(\alpha\) controls the amplification factor and \(b_{\max}\) caps the upper bound. This bias acts as a block-level modality gate, applied between all action queries and tactile keys. Let \(I^c_{qk}\) indicate whether action query \(q\) and tactile key \(k\) belong to the same block (1 if yes, 0 otherwise). The final biased attention logit is:
\begin{equation}
A'_{qk}=A_{qk}+M^{\mathrm{vc}}_{qk}
+I^c_{qk}b_{c(k),k},
\label{eq:biased_attention}
\end{equation}
where \(A\) is the original attention score and \(M^{\mathrm{vc}}\) is the video–tactile masking matrix from Eq.~(10). Note that the bias is computed solely from observed tactile forces, independent of future tactile targets or intermediate denoising predictions.

\subsection{Touch-aware Proxy}
\label{subsec:proxy}

\paragraph{Proxy Construction.}
We employ a differentiable deformation estimator \(\Phi\) as a touch-aware proxy, mapping successive decoded tactile frames to a 3D deformation proxy signal for each sensor. Specifically, for sensor \(s\in\{L,R\}\), let \(I_t^s\) denote the grayscale tactile image and \(\Delta I_t^s=I_t^s-I_{t-1}^s\) its temporal difference (with zero-initialization for the first frame). Using the centered spatial gradients \((g_x^s,g_y^s)=\nabla I_t^s\), we construct a differentiable gradient-aligned deformation field:
\begin{equation}
u_{x,t}^s=\frac{\Delta I_t^s g_x^s}
{\sqrt{(g_x^s)^2+(g_y^s)^2+\epsilon}},\qquad
u_{y,t}^s=\frac{\Delta I_t^s g_y^s}
{\sqrt{(g_x^s)^2+(g_y^s)^2+\epsilon}},
\end{equation}
where \(\epsilon=10^{-6}\). The proxy signal for each sensor is then defined as
\(f_t^s=[\langle u_{x,t}^s\rangle,\langle u_{y,t}^s\rangle,
\langle\partial_xu_{x,t}^s+\partial_yu_{y,t}^s\rangle]\),
where \(\langle\cdot\rangle\) denotes spatial averaging. The first two components summarize tangential deformation, while the third measures the average divergence. Each action chunk corresponds to 8 frames of tactile transition; applying \(\Phi\) to the predicted tactile images yields the predicted proxy signal:
\begin{equation}
\widehat F = \Phi(\widehat o^{\tau,L})\oplus\Phi(\widehat o^{\tau,R})
\in\mathbb R^{C\times 8\times 6},
\label{eq:predicted_proxy}
\end{equation}
where \(C\) is the number of action chunks.

\paragraph{Proxy Supervision.}
Applying the same estimator to the ground-truth tactile trajectories yields the reference signal \(F^*\). We supervise the proxy signal with:
\begin{equation}
\mathcal L_F=0.05\,\operatorname{SmoothL1}(\widehat F,F^*).
\label{eq:force_loss}
\end{equation}
The overall training objective is:
\begin{equation}
\mathcal L=\mathcal L_{\mathrm{video}}+\mathcal L_{\mathrm{action}}
+\lambda_\tau\mathcal{L}_{\mathrm{FM}}^\tau+\mathcal L_F.
\label{eq:total_loss}
\end{equation}
During training, gradients are propagated back to the predicted tactile latents through the frozen decoder and the differentiable estimator \(\Phi\); all Wan VAE parameters remain frozen.

\section{Experiments}

\subsection{Experimental Setup}

We compare with $\pi_{0.5}$, DreamZero, and Naive VT-WAM on eight UniVTAC tasks~\citep{unit2024} (100K training steps and 20 trials per task), nine ManiFeel tasks~\citep{manifeel2025} (60K steps and 50 trials per task, including ablations), and six real-robot conditions (20 trials each); $\pi_{0.5}$ is initialized from released weights, whereas all WAM variants are trained from scratch, and we report success counts and rates under fixed trial budgets.
Additional experimental results and protocol details are provided in Appendix~C.

\subsection{Simulation Results}

As shown in Table~\ref{tab:univtac_results}, pretrained $\pi_{0.5}$ achieves the highest overall rate (43.1\%).
Among models trained from scratch, \method{} reaches 30.0\%, exceeding DreamZero (21.3\%) and the variant without \videoclean{} (26.3\%), with the largest gain on tube insertion ($17/20$ versus $4/20$ for DreamZero and $\pi_{0.5}$).
Failures on HDMI insertion and lifting indicate that the gains remain task dependent.

On ManiFeel, \method{} achieves $147/450$ (32.7\%), improving over DreamZero by 17.1 percentage points and leading on peg insertion, gear insertion, bulb insertion, and bolt-nut assembly (Table~\ref{tab:manifeel_results}).
Although $\pi_{0.5}$ attains a higher overall rate (36.7\%), \method{} performs best on these four contact-dominated tasks.
Across both suites, pretraining provides a strong general task prior, while the proposed tactile pathway is most effective when contact determines the corrective action.

\subsection{Real-Robot Results}

\begin{figure*}[h]
\centering
\includegraphics[width=\textwidth]{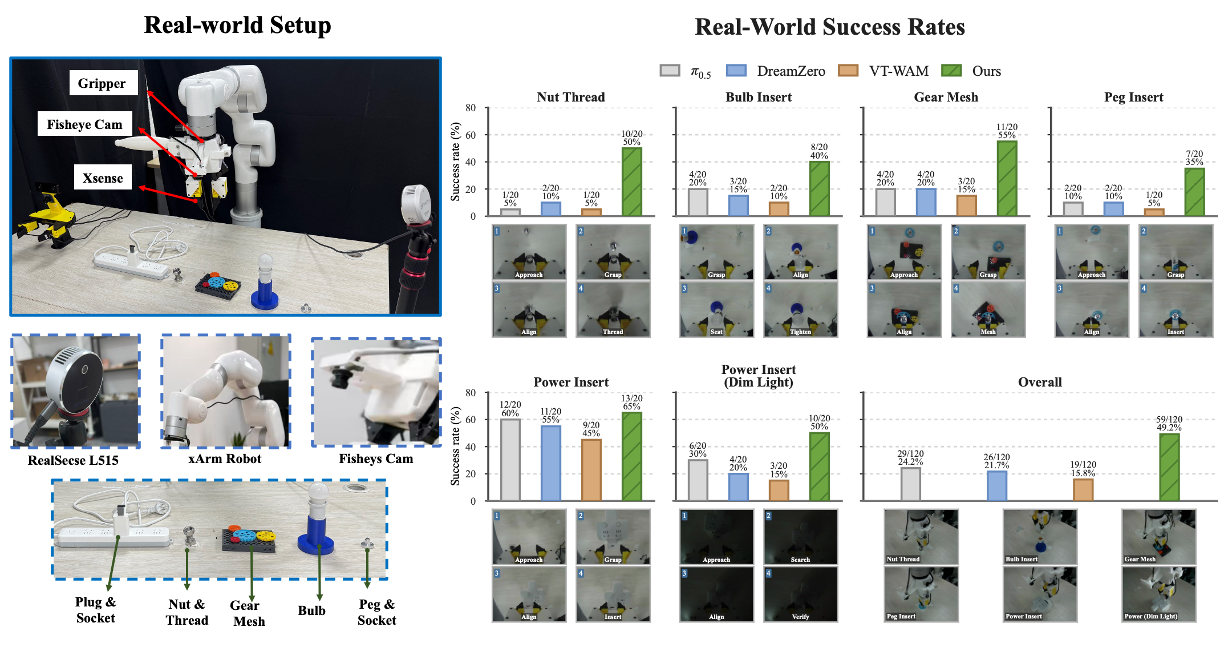}
\caption{Real-robot success rates over six conditions (20 trials each). \method{} ranks first in every condition and retains 50\% success on dim-light power insertion.}
\label{fig:realworld_results}
\end{figure*}

\method{} succeeds in $59/120$ trials (49.2\%), outperforming $\pi_{0.5}$ (24.2\%), DreamZero (21.7\%), and VT-WAM (15.8\%) by 25.0--33.3 percentage points (Figure~\ref{fig:realworld_results}).
It ranks first in all six conditions, including dim-light power insertion, where it retains 50\% success while DreamZero drops from 55\% to 20\%.
This robustness indicates that tactile state complements external vision during post-contact alignment.

\subsection{Tactile Pollution and \videoclean{}}

\begin{table*}[t]
\centering
\footnotesize
\setlength{\tabcolsep}{5pt}
\caption{Agreement with DreamZero at the step-matched 20K checkpoint. Positive paired improvements favor \videoclean{}; all 95\% bootstrap confidence intervals exclude zero.}
\label{tab:videoclean_dreamzero_agreement}
\begin{tabular}{@{}lccccc@{}}
\toprule
Metric & w/o \videoclean{} & \videoclean{} & \videoclean{} change & Paired improvement [95\% CI] & \videoclean{} closer \\
\midrule
MSE $\downarrow$ & 0.001568 & \textbf{0.001227} & $\mathbf{-21.8\%}$ & $+0.000342$ [$0.000117$, $0.000555$] & 84.5\% \\
MAE $\downarrow$ & 0.020904 & \textbf{0.017190} & $\mathbf{-17.8\%}$ & $+0.003714$ [$0.002877$, $0.004591$] & 90.5\% \\
PSNR $\uparrow$ & 31.016 & \textbf{32.020} & $\mathbf{+1.005}$ dB & $+1.005$ [$0.646$, $1.356$] & 81.0\% \\
SSIM $\uparrow$ & 0.98440 & \textbf{0.98683} & $\mathbf{+0.00243}$ & $+0.00243$ [$0.00005$, $0.00468$] & 70.2\% \\
\bottomrule
\end{tabular}
\end{table*}

\begin{table}[t]
\centering
\scriptsize
\setlength{\tabcolsep}{3.0pt}
\caption{Prediction-to-DreamZero MSE across four rollout chunks at 20K. The reduction from \videoclean{} reaches 38.5\% at H4.}
\label{tab:videoclean_horizon}
\resizebox{\columnwidth}{!}{
\begin{tabular}{@{}lcccc@{}}
\toprule
Chunk & w/o \videoclean{} & \videoclean{} & Error reduction & \videoclean{} closer \\
\midrule
H1 & \textbf{0.000695} & 0.000706 & $-1.7\%$ & 13.1\% \\
H2 & 0.001395 & \textbf{0.001349} & $+3.3\%$ & 70.2\% \\
H3 & 0.001735 & \textbf{0.001346} & $\mathbf{+22.4\%}$ & 88.1\% \\
H4 & 0.002448 & \textbf{0.001505} & $\mathbf{+38.5\%}$ & 90.5\% \\
\bottomrule
\end{tabular}}
\end{table}

We test whether \videoclean{} preserves DreamZero's visual dynamics using 84 paired held-out UniVTAC samples at step-matched 20K checkpoints, separate from the 100K control evaluation in Table~\ref{tab:univtac_results}.
Both tactile variants use identical samples, conditioning frames, and diffusion seeds, while the 20K DreamZero checkpoint serves as the reference.
Their paired prediction-to-prediction distance measures how much tactile conditioning perturbs the learned visual trajectory.

\videoclean{} reduces prediction-to-DreamZero MSE and MAE by 21.8\% and 17.8\%, respectively, and is closer to DreamZero on 84.5\% and 90.5\% of samples under these metrics (Table~\ref{tab:videoclean_dreamzero_agreement}).
The paired confidence intervals support more reliable preservation of DreamZero's visual dynamics than unrestricted tactile fusion.

\begin{figure}[t]
\centering
\includegraphics[width=\columnwidth]{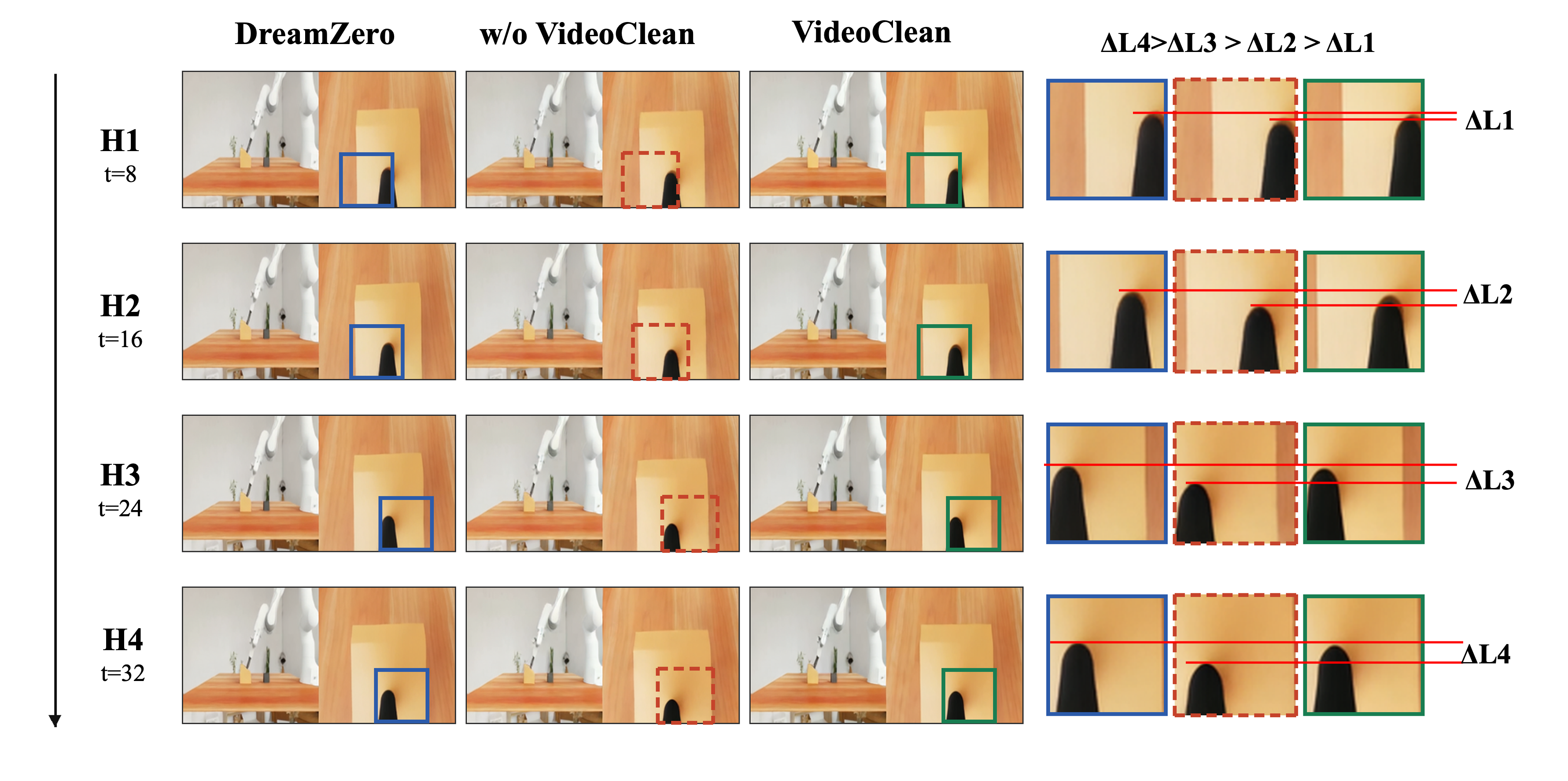}
\caption{Step-matched 20K rollouts from H1 (top) to H4 (bottom). Blue, red dashed, and green boxes denote DreamZero, w/o \videoclean{}, and \videoclean{}; the right crops enlarge the same region. \videoclean{} better preserves the DreamZero trajectory at long horizons.}
\label{fig:videoclean_rollout}
\end{figure}

\videoclean{} is comparable to unrestricted fusion at H1 but becomes increasingly effective as autoregressive error accumulates, reducing MSE by 3.3\%, 22.4\%, and 38.5\% at H2--H4 (Table~\ref{tab:videoclean_horizon}; Figure~\ref{fig:videoclean_rollout}).
At H4, it is closer to DreamZero on 90.5\% of samples, showing that its main effect is to suppress compounding tactile-induced visual drift.

\begin{table}[t]
\centering
\scriptsize
\setlength{\tabcolsep}{2.4pt}
\caption{Ground-truth prediction quality at the step-matched 20K checkpoint. Positive $\Delta$ favors \videoclean{}; all paired 95\% confidence intervals include zero.}
\label{tab:videoclean_ground_truth}
\resizebox{\columnwidth}{!}{
\begin{tabular}{@{}lrrl@{}}
\toprule
Metric & w/o \videoclean{} & \videoclean{} & Paired $\Delta$ [95\% CI] \\
\midrule
MSE $\downarrow$ & \textbf{0.001748} & 0.001777 & $-0.000029$ [$-0.000257$, $0.000202$] \\
MAE $\downarrow$ & 0.021034 & \textbf{0.020976} & $+0.000058$ [$-0.000765$, $0.000915$] \\
PSNR $\uparrow$ & 30.111 & \textbf{30.305} & $+0.194$ [$-0.078$, $0.469$] \\
SSIM $\uparrow$ & 0.98169 & \textbf{0.98231} & $+0.00062$ [$-0.00134$, $0.00255$] \\
\bottomrule
\end{tabular}}
\end{table}

All confidence intervals in Table~\ref{tab:videoclean_ground_truth} include zero, indicating no statistically detectable change in ground-truth prediction quality.
Thus, \videoclean{} protects the visual prior and limits long-horizon drift, while virtual-force supervision and contact-aware routing determine how tactile representations guide action generation.

\subsection{Effect of Contact-Aware Attention Bias}

\begin{figure}[h]
\centering
\includegraphics[width=\columnwidth]{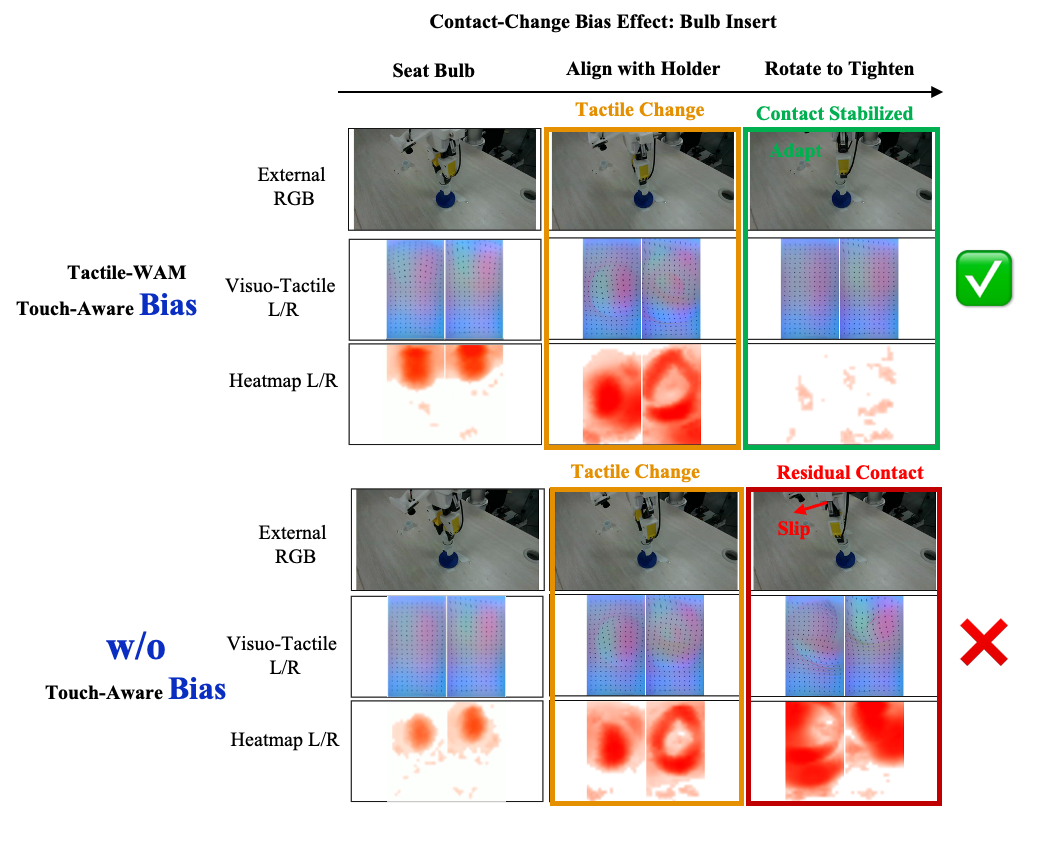}
\caption{Contact-aware attention during bulb insertion. With the bias (top), tactile change redirects action attention and stabilizes contact; without it (bottom), residual contact leads to slip.}
\label{fig:bias_realworld}
\end{figure}

During bulb insertion, the bias redirects action attention toward tactile keys after initial contact, enabling correction before the error is visible in external RGB and weakening as contact stabilizes (Figure~\ref{fig:bias_realworld}).
Without the bias, residual contact develops into slip.
This comparison supports the proposed mechanism but, without a bias-only real-robot variant, does not isolate its success-rate contribution.

\subsection{Ablation Study}

\begin{table}[t]
\centering
\small
\setlength{\tabcolsep}{3.5pt}
\caption{Cumulative ablation on ManiFeel at 60K steps. Force denotes virtual-force auxiliary supervision; Bias denotes the contact-aware tactile attention bias.}
\label{tab:manifeel_ablation}
\resizebox{\columnwidth}{!}{
\begin{tabular}{lccccc}
\toprule
Method & Future Touch & \videoclean{} & Force & Bias & Success \\
\midrule
RGB-only WAM (DreamZero) & -- & -- & -- & -- & 0.156 \\
Naive VT-WAM & \cmark & -- & -- & -- & 0.151 \\
$+\videoclean{}$ & \cmark & \cmark & -- & -- & 0.284 \\
Full \method{} & \cmark & \cmark & \cmark & \cmark & \textbf{0.327} \\
\bottomrule
\end{tabular}}
\end{table}

\paragraph{Naive future tactile prediction.}
Naive VT-WAM reaches 15.1\%, slightly below DreamZero's 15.6\%, suggesting that direct tactile injection may interfere with the shared representation.

\paragraph{Effect of \videoclean{}.}
Adding \videoclean{} raises success from 15.1\% to 28.4\% (+13.3 percentage points), the largest ablation gain, indicating that visual--tactile interaction is a greater bottleneck than tactile availability alone.

\paragraph{Virtual-force supervision and contact-aware bias.}
Virtual-force supervision and contact-aware bias jointly raise success from 28.4\% to 32.7\% (+4.3 percentage points).
The two components target representation content and attention timing, respectively.

\paragraph{Overall gain and identification limit.}
Full \method{} improves over DreamZero from 15.6\% to 32.7\% (+17.1 percentage points; $2.10\times$).
Because force supervision and attention bias are introduced together, the ablation measures only their joint contribution; their individual effects require force-only and bias-only variants.

\section{Conclusion}

We presented Tactile-WAM, a tactile-aware world action model that addresses tactile pollution via asymmetric attention and deformation-based proxy. It achieves state-of-the-art results on UniVTAC and ManiFeel, with 32.7\% success and a 17.1-point improvement over DreamZero, while reducing visual trajectory MSE by 21.8\%. Physical evaluation ranks first across all conditions. Future work will investigate individual component effects and generalizability.

\clearpage

\bibliography{references}
\clearpage
\appendix

\setcounter{section}{0}
\renewcommand{\thesection}{\Alph{section}}

\setcounter{figure}{0}
\renewcommand{\thefigure}{S\arabic{figure}}

\setcounter{table}{0}
\renewcommand{\thetable}{S\arabic{table}}

\section*{Supplementary Material}
\addcontentsline{toc}{section}{Supplementary Material}

\section{Related Work}
\label{Related_Work}

\paragraph{World action models.}
Joint observation-action denoising connects generative video priors to control~\citep{pad2024,vpp2024,dreamzero2026}, while action-conditioned foresight compares predicted and observed scenes for failure recovery~\citep{pan2025self}.
X-WAM adds multi-view RGB-D prediction and 4D reconstruction to Wan2.2~\citep{xwam2026}.
WorldAgen uses separated heads and a mixed unidirectional mask for state-action prediction and test-time adaptation~\citep{wan2026worldagen}; LingBot-VA controls interaction through modality-specific transformer pathways~\citep{lingbotva2026}.
H-GAR couples predicted observations with hierarchical action refinement~\citep{hgar2026}.
\method{} differs by predicting touch as a physical state and constraining its interaction with video and action tokens.

\paragraph{Touch for manipulation and prediction.}
Optical tactile sensing exposes local contact geometry and deformation that may be weakly observable in RGB~\citep{gelsight2017,digit2020,tactip2018}.
UniT and Sparsh learn transferable representations~\citep{unit2024,sparsh2024}; recent work studies tactile reasoning, adaptive visual-tactile fusion, and cross-modal alignment~\citep{stola2026,touchformer2026,tlvalign2026}.
Reactive and multi-fingered policies use current touch for correction~\citep{rdp2025,jia2025multifingered}, while force-aware learning targets gentle manipulation~\citep{gentle2026}.
ManiFeel provides contact-rich tasks and matched evaluation~\citep{manifeel2025}, and visuo-tactile world models predict future touch for control~\citep{vtwm2026}.
\method{} instead encodes touch with the visual Wan VAE and supervises decoded tactile latents.

\paragraph{Cross-modal interference and tactile pollution.}
LingBot-VA prevents modality-specific representation interference with separately parameterized video and action transformer pathways~\citep{lingbotva2026}.
Our setting instead concatenates video, tactile, action, and state tokens inside one Wan-based DiT.
\videoclean{} therefore removes tactile-key access only for video queries, leaving action-to-tactile attention intact; \emph{tactile pollution} names the failure in which sparse tactile events perturb the pretrained visual path.

\paragraph{Auxiliary supervision.}
FoAM predicts action consequences, ReconVLA reconstructs task-relevant visual regions, and ForeDiffusion aligns a policy with constructed future views~\citep{foam2026,reconvla2026,forediffusion2026}.
LingBot-VA 2.0 extends this idea over longer visual horizons with training-only heads~\citep{lingbotva2_2026}, while MP1 improves generalization through a loss that does not slow inference~\citep{mp1_2026}.
\method{} applies the same training-time principle to virtual force derived from decoded tactile futures.

\section{Model Details}
\label{sec:model_details}

This section describes the architecture of \method{}, including its shared
latent representation, multimodal tokenization, causal attention structure,
touch-aware routing, prediction heads, and optimization configuration.

\subsection{Backbone and Latent Representation}

\method{} is built on the Wan2.2-TI2V-5B diffusion Transformer
backbone~\citep{wan2025}.  The backbone contains 30 Transformer blocks with a
hidden dimension of 3072, 24 attention heads, and a feed-forward dimension of
14336.  Each feed-forward block follows
\(\mathrm{Linear}(3072,14336)\)--GELU--
\(\mathrm{Linear}(14336,3072)\).

RGB and optical tactile observations share the frozen Wan2.2 VAE.  The VAE
maps both modalities to a 48-channel latent space with spatial and temporal
downsampling factors of 16 and 4, respectively.  For the two tactile sensors,
we form a horizontal mosaic
\begin{equation}
\mathcal M(o^{\tau,L},o^{\tau,R})
=
[\,o^{\tau,L}\mid o^{\tau,R}\,],
\label{eq:appendix_tactile_mosaic}
\end{equation}
and resize the mosaic to the VAE input size.  Sharing the VAE places visual
and tactile observations in a common generative latent space without
introducing a separate tactile tokenizer.

Both RGB sequences and complete tactile mosaics are resized bilinearly to
\(320\times160\) before VAE encoding.  Integer-valued images are first divided
by 255 and then normalized with channel-wise mean and standard deviation 0.5,
giving the VAE input range \([-1,1]\).  Resizing is applied to the complete
left-right mosaic rather than independently to the two sensor images.

\subsection{Multimodal Tokenization}

Video latents use Wan's native three-dimensional patchification with patch
size \((1,2,2)\).  Tactile latents are unfolded into \(2\times2\) spatial
patches.  Each tactile patch therefore contains
\(48\times2\times2=192\) values and is projected to the 3072-dimensional DiT
space by a learnable linear layer.  A learnable tactile-modality embedding is
added after projection.

An action \(a\in\mathbb R^{D_a}\) is first mapped to 3072 dimensions and
combined with a sinusoidal diffusion-time embedding through
\begin{equation}
\mathrm{Linear}(6144,3072)
\rightarrow \mathrm{SiLU}
\rightarrow \mathrm{Linear}(3072,3072).
\label{eq:appendix_action_projection}
\end{equation}
Proprioceptive states are projected by a two-layer MLP.  Language instructions
are encoded by UMT5-XXL into 4096-dimensional token features, projected to
3072 dimensions, and injected into each DiT block through cross-attention.

Video tokens use the native three-dimensional rotary positional encoding
(3-D RoPE), while tactile, action, and state tokens use one-dimensional RoPE.
Within each modality, tokens are ordered by causal block.  Their conceptual
layout is
\begin{verbatim}
[condition and noisy video tokens]
[clean tactile-history tokens]
[noisy future-tactile tokens]
[action tokens]
[state tokens]
\end{verbatim}
RoPE and the causal block mask preserve temporal correspondence across these
modality groups.

\subsection{Temporal Block Organization}

The model receives 33 image frames, which the temporal VAE compresses to nine
latent frames.  The first latent frame is retained as a clean condition; the
remaining eight latent frames are divided into four non-overlapping causal
blocks, with two latent frames per block.  Each block is aligned with a
24-step action chunk and one proprioceptive state token.  Visual, tactile,
action, and state tokens are stored in modality-wise groups rather than
interleaved frame by frame.  A block identifier attached to every token
provides the temporal alignment used by RoPE and the attention mask.

Actions and states use a shared width of eight dimensions.  Lower-dimensional
inputs are zero-padded and accompanied by validity masks.  Each valid
dimension \(x_d\) is normalized with its first and 99th percentiles:
\begin{equation}
\widetilde{x}_d
=
\operatorname{clip}\!\left(
2\frac{x_d-q_{01,d}}{q_{99,d}-q_{01,d}}-1,
-1,1
\right).
\label{eq:appendix_action_normalization}
\end{equation}
Dimensions with zero percentile range retain their original value.  Padding
dimensions are excluded from the action loss.

\subsection{Causal Multimodal Attention}

Let \(M^{\mathrm{causal}}\) denote the native block-causal attention mask.
\videoclean{} adds a directional mask that prevents visual queries from
reading tactile keys:
\begin{equation}
M^{\mathrm{vc}}_{qk}
=
\begin{cases}
-\infty, & G(q)=V\ \wedge\ G(k)=T,\\
0, & \text{otherwise},
\end{cases}
\label{eq:appendix_videoclean}
\end{equation}
where \(G(\cdot)\) identifies the modality of a token.  Action queries retain
access to tactile keys, and tactile queries preserve their causally permitted
multimodal context.  The mask therefore isolates only the pathway that can
inject tactile information into visual prediction.

The touch-aware bias selectively strengthens action-to-tactile attention.
For causal block \(c\), an observed tactile pair is mapped to a
six-dimensional proxy \(F_c^{\mathrm{obs}}\).  We convert its magnitude into
a bounded block-level score
\begin{equation}
b_c
=
\operatorname{clip}\!\left[
\alpha\tanh\!\left(
\frac{\operatorname{ReLU}
(\lVert F_c^{\mathrm{obs}}\rVert_2-\theta)}{T}
\right),
0,b_{\max}
\right].
\label{eq:appendix_touch_bias}
\end{equation}
The complete attention logit is
\begin{equation}
A'_{qk}
=
\frac{q^\top k}{\sqrt d}
+M^{\mathrm{causal}}_{qk}
+M^{\mathrm{vc}}_{qk}
+B^\tau_{qk},
\label{eq:appendix_attention_logit}
\end{equation}
where \(B^\tau_{qk}=b_c\) only for an action query and a tactile key in the
same causally accessible block.  Because the bias modifies only connections
already admitted by \(M^{\mathrm{causal}}\), it cannot reveal future tokens.

\subsection{Touch-Aware Proxy}

The fixed differentiable operator \(\Phi\) summarizes deformation between two
tactile frames.  RGB inputs are mapped to grayscale using
\(0.299R+0.587G+0.114B\), after conversion to \([0,1]\).  Let
\(\Delta I=I_t-I_{t-1}\).  Spatial derivatives use the centered
one-dimensional kernels
\(k_x=\tfrac12[-1,0,1]\) and \(k_y=k_x^\top\), with one-pixel zero padding.
Writing the resulting gradients as \(g_x\) and \(g_y\), the
gradient-aligned deformation field is
\begin{equation}
u_x
=
\frac{\Delta I\,g_x}
{\sqrt{g_x^2+g_y^2+\epsilon}},
\qquad
u_y
=
\frac{\Delta I\,g_y}
{\sqrt{g_x^2+g_y^2+\epsilon}},
\label{eq:appendix_deformation_field}
\end{equation}
where \(\epsilon>0\) is a numerical stabilizer.  Each sensor produces
\begin{equation}
f^\tau
=
[\langle u_x\rangle,\langle u_y\rangle,
\langle\partial_xu_x+\partial_yu_y\rangle],
\label{eq:appendix_single_sensor_proxy}
\end{equation}
where \(\langle\cdot\rangle\) denotes spatial averaging.  Concatenating the
left and right outputs gives
\begin{equation}
F^\tau
=
[\langle u_x^L\rangle,\langle u_y^L\rangle,
\langle\mathrm{div}^L\rangle,
\langle u_x^R\rangle,\langle u_y^R\rangle,
\langle\mathrm{div}^R\rangle]\in\mathbb R^6.
\label{eq:appendix_proxy_order}
\end{equation}

The proxy is used at two temporal resolutions.  For the auxiliary target,
each causal block contains eight consecutive frame-to-frame transitions:
\begin{equation}
F_c^*
=
\big[
\Phi(I_{u_c},I_{u_c+1}),\ldots,
\Phi(I_{u_c+7},I_{u_c+8})
\big]\in\mathbb R^{8\times6}.
\label{eq:appendix_proxy_sequence}
\end{equation}
Across \(C\) blocks, the target tensor therefore has shape
\(\mathbb R^{C\times8\times6}\), ordered as
\((\text{block},\text{transition},\text{feature})\).  In contrast, the
observed cue used by the attention bias is a single non-adjacent difference,
\begin{equation}
F_c^{\mathrm{obs}}=\Phi(I_{u_c-8},I_{u_c}),
\label{eq:appendix_observed_proxy}
\end{equation}
and is not a sum of eight differences.

The operator \(\Phi\) has no learnable parameters.  For tactile prediction,
the proxy loss is computed from decoded tactile latents and remains
differentiable with respect to the predicted latents.  The VAE parameters stay
frozen.  For attention routing, the observed proxy is detached before the
bias is constructed.

\subsection{Prediction Heads and Objective}

The tactile output head applies
\(\mathrm{Linear}(3072,192)\) to each tactile token and unpatchifies the result
into a 48-channel latent flow field.  The action head is a two-layer MLP,
\begin{equation}
\mathrm{Linear}(3072,1024)
\rightarrow \mathrm{ReLU}
\rightarrow \mathrm{Linear}(1024,D_a).
\label{eq:appendix_action_head}
\end{equation}
The shared implementation width is \(D_a=8\), with invalid padded dimensions
masked as described above.
The model jointly predicts visual latents, tactile latents, and action tokens.
Its objective is
\begin{equation}
\mathcal L
=
\mathcal L_{\mathrm{video}}
+\mathcal L_{\mathrm{action}}
+\lambda_\tau\mathcal L_{\mathrm{FM}}^\tau
+\lambda_{\mathrm{proxy}}\mathcal L_{\mathrm{proxy}},
\label{eq:appendix_model_objective}
\end{equation}
where
\(\mathcal L_{\mathrm{proxy}}
=\operatorname{SmoothL1}(\widehat F,F^*)\).
Each modality-specific loss is normalized over its own feature dimensions
before the terms are combined.

\subsection{Training Configuration}

\paragraph{Optimization.}
We fully fine-tune the DiT backbone together with the multimodal projections,
modality embeddings, modified attention layers, and prediction heads.  The
pretrained VAE and language encoder remain frozen and are kept in inference
mode.  All trainable parameters are optimized jointly with AdamW using a peak
learning rate of \(1\times10^{-5}\), \(\beta_1=0.95\), \(\beta_2=0.999\), and
\(\epsilon=10^{-8}\).  Weight decay is \(1\times10^{-5}\) for eligible weight
matrices and zero for bias and normalization parameters.  The learning rate
is linearly warmed up during the first \(5\%\) of optimization and then
decayed with a cosine schedule.

\paragraph{Flow-matching construction.}
The scheduler contains 1000 discrete training noise levels.  For a clean
latent or action target \(x_0\), we sample
\(\epsilon\sim\mathcal N(0,I)\) and construct
\begin{equation}
x_t=(1-\sigma_t)x_0+\sigma_t\epsilon,
\qquad
v_t^*=\epsilon-x_0.
\label{eq:appendix_flow_training}
\end{equation}
One noise level is sampled uniformly for each causal block.  Frames within
the same block share this level, and the temporally aligned visual, tactile,
and action targets use the same block-level timestep.  The first conditioning
frame remains clean.  The video, tactile, and action branches minimize
timestep-weighted mean-squared errors on \(v_t^*\), with the scheduler weights
normalized to unit mean.

\paragraph{Loss balancing and numerical stability.}
The modality coefficients in Eq.~\eqref{eq:appendix_model_objective} are kept
fixed throughout optimization.  Action padding and unavailable targets are
masked before reduction.  Each branch is averaged over its valid tokens and
feature dimensions before the weighted losses are added.  We use
BF16 mixed precision, enable TF32 matrix multiplications and activation
recomputation, and clip the global gradient norm to 1.0.  The random seed is
set to 42.

\paragraph{Training scale and hardware.}
Training uses eight NVIDIA A100 GPUs with a per-device micro-batch size of 1
and one gradient-accumulation step, giving an effective global batch size of
8.  Full-parameter optimization is distributed with DeepSpeed ZeRO-2; optimizer
states and gradients are sharded across devices, without CPU optimizer
offloading.

\section{Experiments Details}

\subsection{Experimental Setup}

\paragraph{Simulation protocols.}
We evaluate on UniVTAC~\citep{unit2024} and ManiFeel~\citep{manifeel2025}.
UniVTAC contains eight manipulation tasks; each method is trained for 100K steps and evaluated for 20 trials per task, giving 160 trials per method.
ManiFeel contains nine tasks; each model is trained for 60K steps and evaluated for 50 trials per task, giving 450 trials per method.
The ablation study is conducted only on ManiFeel with the same 60K-step budget.
For ManiFeel, observations are sampled at 10 fps, the simulator runs at 60 Hz with control decimation 4, and the policy predicts 24 actions and executes 8 actions before replanning.
Six-dimensional task actions are zero-padded to the model's 7-D interface and unpadded before execution.

\paragraph{Real-robot protocol.}
The physical evaluation covers nut threading, bulb insertion, gear meshing, peg insertion, and power insertion.
Power insertion is additionally evaluated under dim lighting, yielding six conditions.
Each method performs 20 trials per condition and 120 trials in total.
Hardware, sensor installation, data collection, control frequency, task initialization, success criteria, and reset procedures are provided in the supplementary material.

\paragraph{Baselines and training fairness.}
We compare with $\pi_{0.5}$, DreamZero, and Naive VT-WAM.
$\pi_{0.5}$ is fine-tuned from its released base weights and therefore benefits from pretrained initialization, whereas DreamZero, Naive VT-WAM, and our variants are trained from scratch in each environment.
We report all absolute results, but comparisons among the from-scratch WAM variants provide the most controlled measure of the contribution of tactile modeling.
Success rate is the primary control metric; every table reports the number of successful executions over the fixed trial count.

\subsection{Qualitative Execution Trajectories}
\label{sec:qualitative_trajectories}

This section presents successful task-level execution trajectories for all
simulation and real-robot tasks evaluated in the paper. Each visualization
preserves the temporal ordering, task-specific stages, synchronized camera
views, and tactile observations used during execution.

\subsubsection{Qualitative Results on Simulation Tasks}
\label{sec:sim_vis}

Figures~\ref{fig:sim_manifeel_ball_sorting}--\ref{fig:sim_univtac_bottle_shelving}
present successful execution trajectories from all simulation tasks.
Figures~\ref{fig:sim_manifeel_ball_sorting}--\ref{fig:sim_manifeel_usb_insertion}
cover the nine ManiFeel tasks, with front, side, wrist, and bilateral
visuo-tactile observations. Figures~\ref{fig:sim_univtac_grasp_classify}--
\ref{fig:sim_univtac_bottle_shelving} cover the eight UniVTac tasks, with
external, wrist, and bilateral visuo-tactile observations.

\subsubsection{Qualitative Results on Real-Robot Experiments}
\label{sec:real_vis}

Figures~\ref{fig:real_bulb_insertion}--\ref{fig:real_power_insertion}
illustrate successful execution sequences from all seven real-world robotic
experiments. Each visualization synchronizes the external RGB view, wrist RGB
view, bilateral visuo-tactile images, and tactile heatmaps.

\newcommand{\QualitativeTrajectory}[4]{%
\begin{figure}[!htbp]
    \centering
    \includegraphics[width=#1\textwidth,keepaspectratio]{#2}
    \caption{#3}
    \label{#4}
\end{figure}%
}
\onecolumn

\QualitativeTrajectory{0.94}
{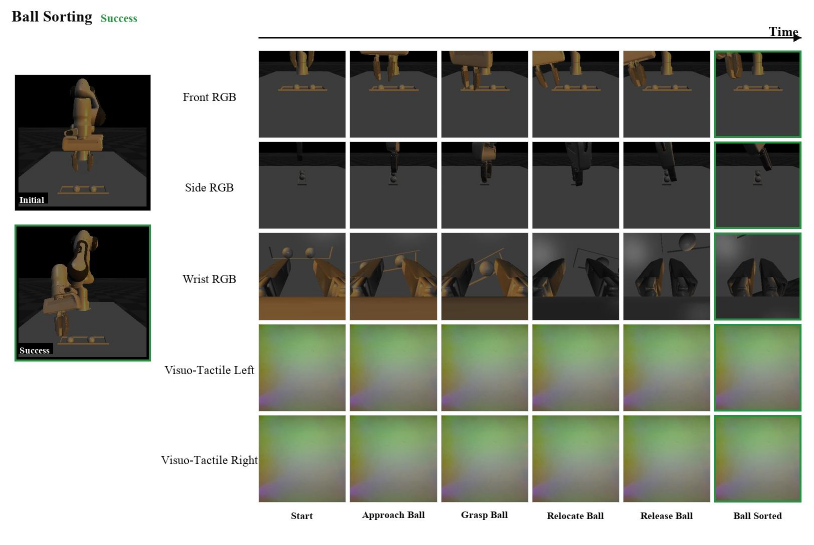}
{Successful ManiFeel execution for ball sorting. The robot approaches and
grasps the ball, relocates it toward the target region, and releases it at the
desired location.}
{fig:sim_manifeel_ball_sorting}

\QualitativeTrajectory{0.94}
{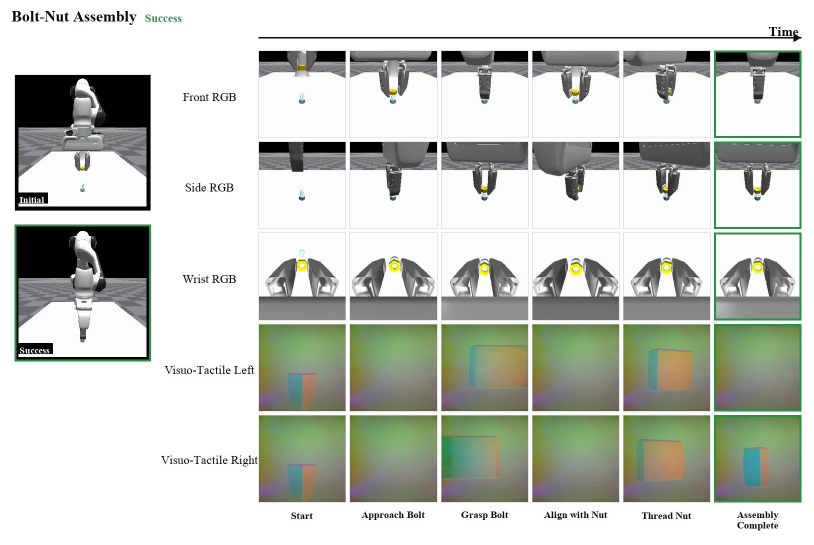}
{Successful ManiFeel execution for bolt--nut assembly. The sequence shows
approach, grasp, contact establishment, axis alignment, and threading.}
{fig:sim_manifeel_bolt_nut}

\QualitativeTrajectory{0.94}
{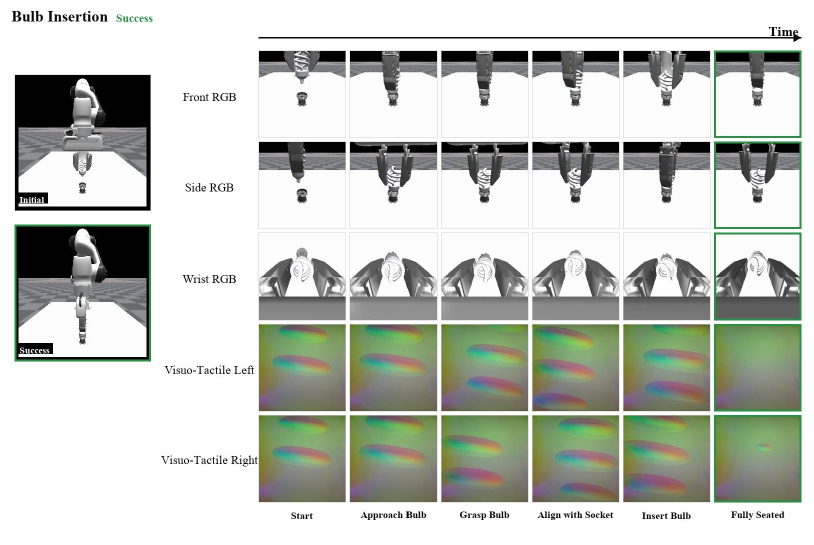}
{Successful ManiFeel execution for bulb insertion. The bulb is grasped,
aligned with the socket, inserted, and seated while tactile observations
capture the evolving contact state.}
{fig:sim_manifeel_bulb_insertion}

\QualitativeTrajectory{0.94}
{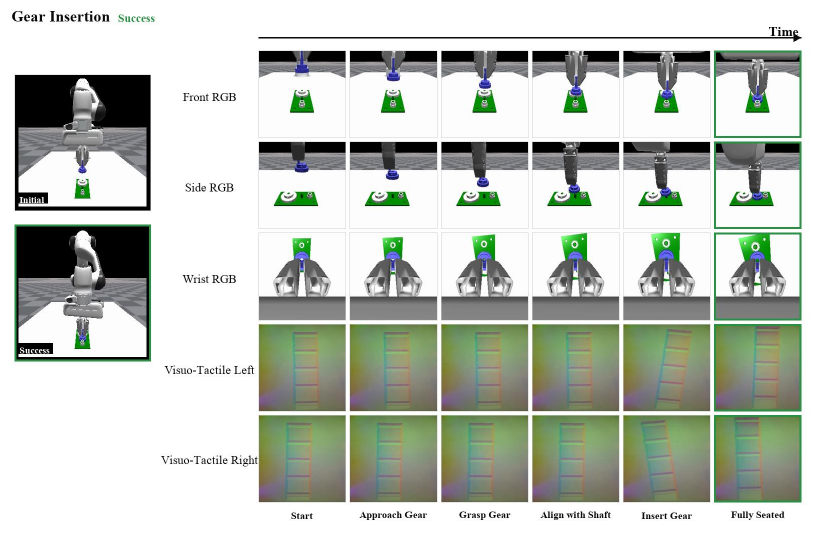}
{Successful ManiFeel execution for gear insertion, including approach, grasp,
alignment with the shaft, insertion, and final seating.}
{fig:sim_manifeel_gear_insertion}

\QualitativeTrajectory{0.94}
{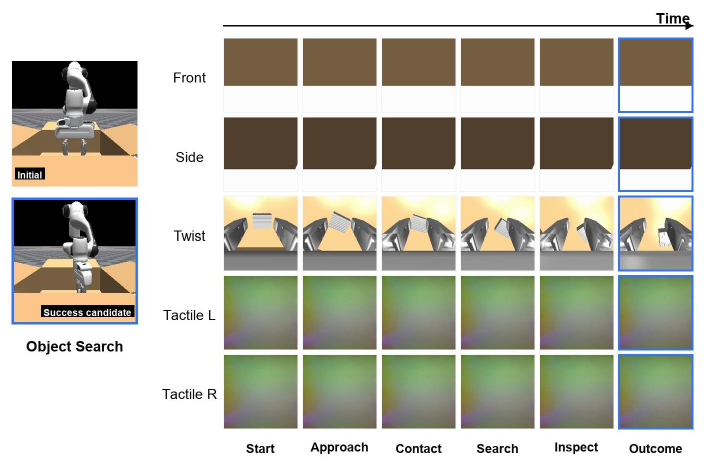}
{Successful ManiFeel execution for object search. The robot explores the
workspace, establishes tactile contact, localizes the target, and completes
the search.}
{fig:sim_manifeel_object_search}

\QualitativeTrajectory{0.94}
{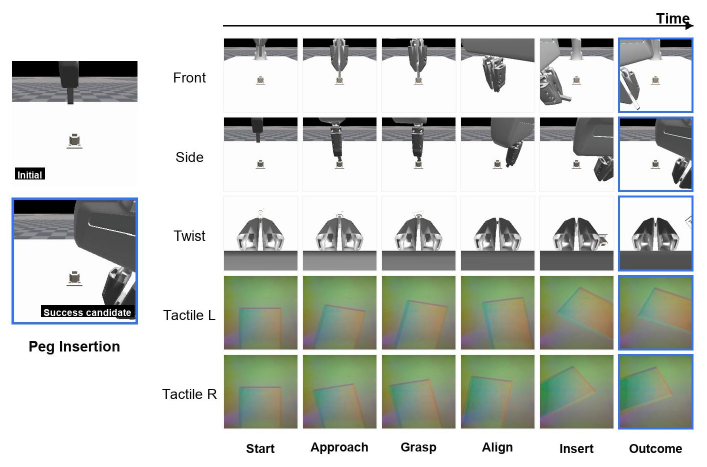}
{Successful ManiFeel execution for peg insertion. The robot grasps the peg,
aligns it with the opening through contact, and completes insertion.}
{fig:sim_manifeel_peg_insertion}

\QualitativeTrajectory{0.94}
{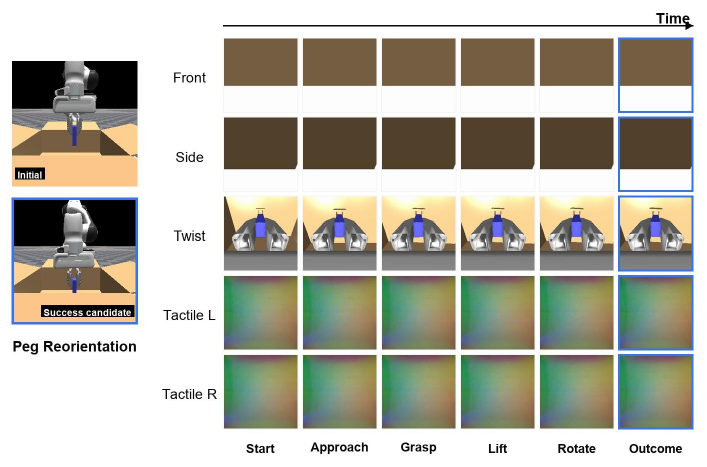}
{Successful ManiFeel execution for peg reorientation. The peg is grasped,
rotated under contact, and placed in the required final orientation.}
{fig:sim_manifeel_peg_reorientation}

\QualitativeTrajectory{0.94}
{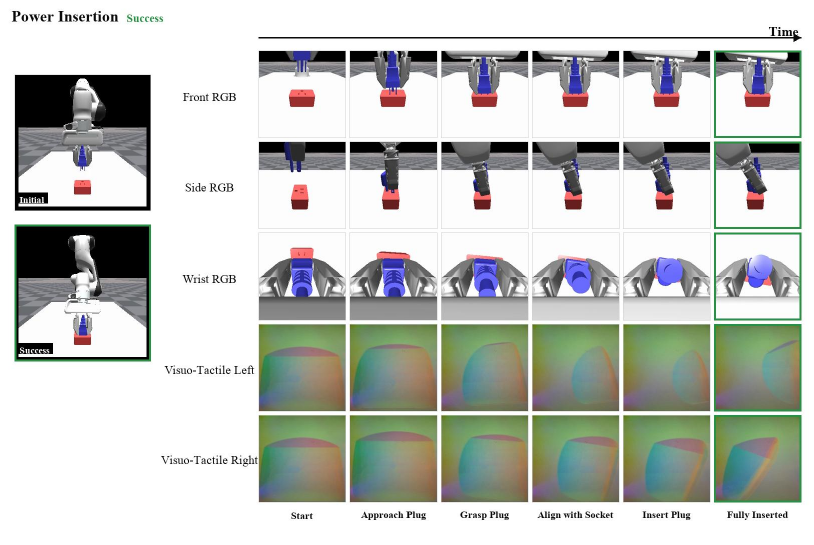}
{Successful ManiFeel execution for power-plug insertion. The robot approaches
the plug, grasps it, aligns it with the socket, and completes insertion.}
{fig:sim_manifeel_power_insertion}

\QualitativeTrajectory{0.94}
{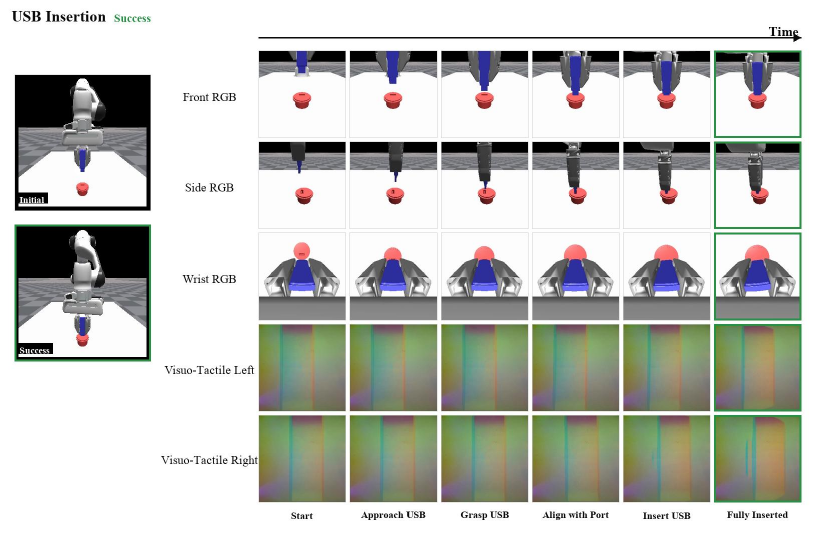}
{Successful ManiFeel execution for USB insertion, showing approach, grasp,
port alignment, insertion, and full seating.}
{fig:sim_manifeel_usb_insertion}

\clearpage

\QualitativeTrajectory{0.94}
{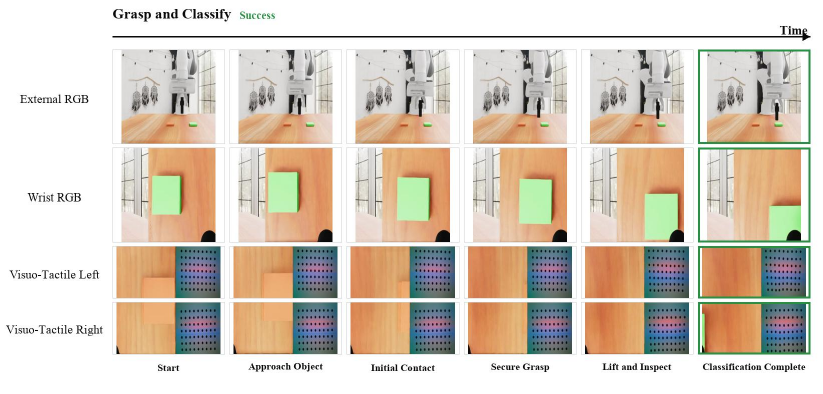}
{Successful UniVTac execution for grasp and classify. The robot establishes a
secure grasp, lifts and inspects the object, and maintains stable contact
through classification.}
{fig:sim_univtac_grasp_classify}

\QualitativeTrajectory{0.94}
{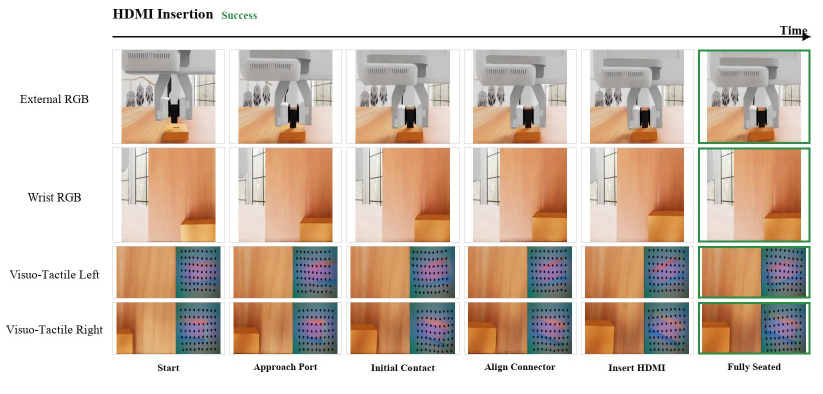}
{Successful UniVTac execution for HDMI insertion. The connector is grasped,
aligned with the port, inserted, and fully seated.}
{fig:sim_univtac_hdmi_insertion}

\QualitativeTrajectory{0.94}
{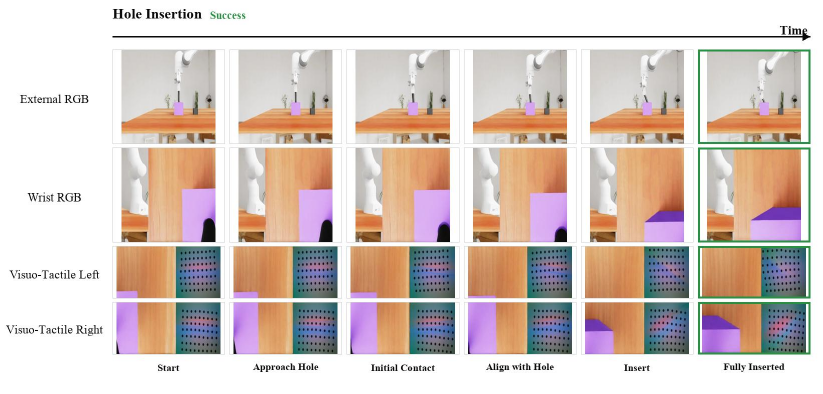}
{Successful UniVTac execution for hole insertion. Contact-guided alignment is
followed by insertion to the target depth.}
{fig:sim_univtac_hole_insertion}

\QualitativeTrajectory{0.94}
{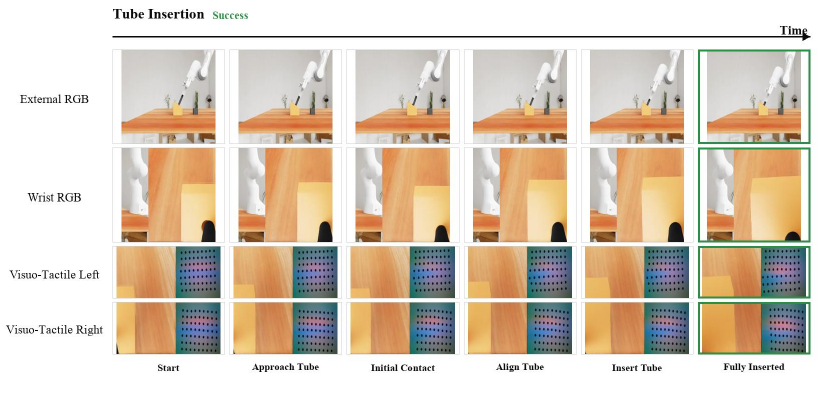}
{Successful UniVTac execution for tube insertion, including approach, grasp,
coaxial alignment, insertion, and final seating.}
{fig:sim_univtac_tube_insertion}

\QualitativeTrajectory{0.94}
{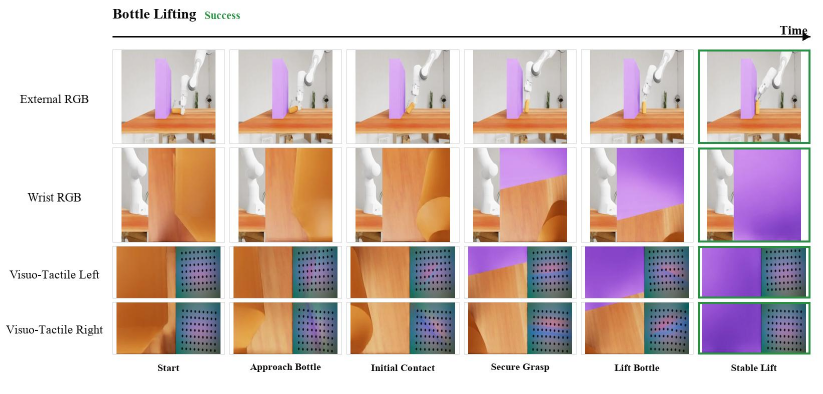}
{Successful UniVTac execution for bottle lifting. The robot forms a stable
grasp and maintains contact throughout the lift.}
{fig:sim_univtac_bottle_lifting}

\QualitativeTrajectory{0.94}
{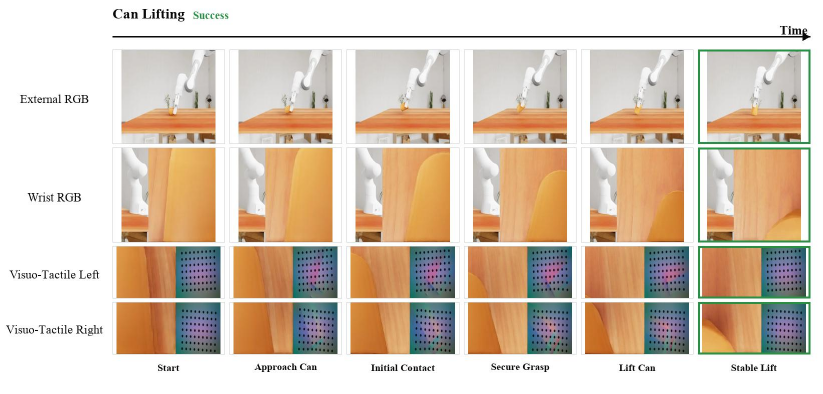}
{Successful UniVTac execution for can lifting. The can is securely grasped and
lifted without slip.}
{fig:sim_univtac_can_lifting}

\QualitativeTrajectory{0.94}
{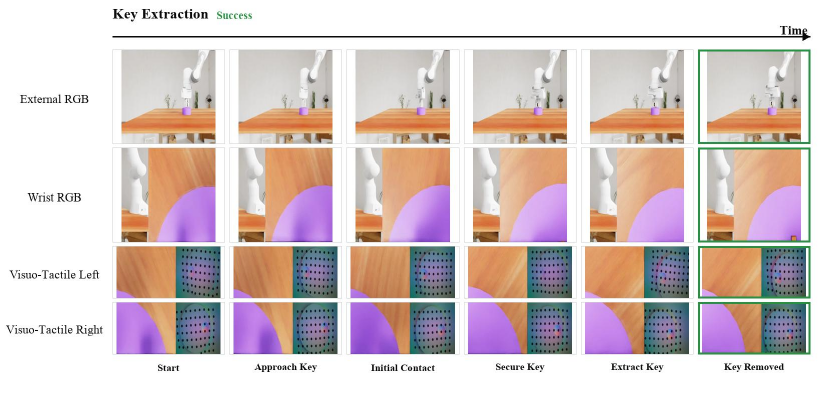}
{Successful UniVTac execution for key extraction. The robot grasps the key,
maintains alignment under contact, and removes it completely.}
{fig:sim_univtac_key_extraction}

\QualitativeTrajectory{0.94}
{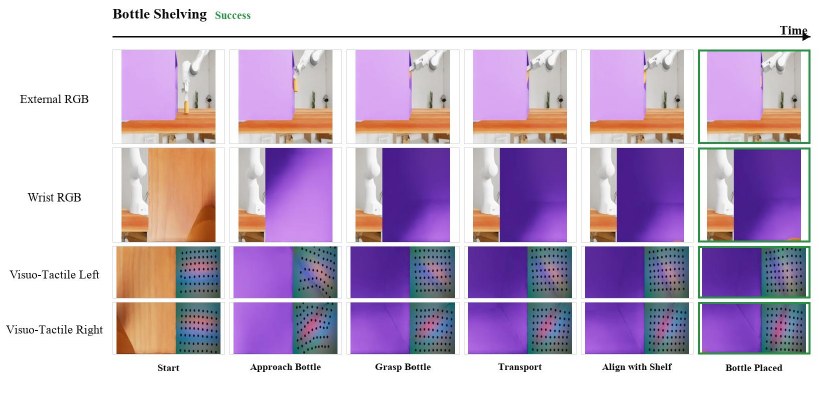}
{Successful UniVTac execution for bottle shelving. The robot grasps and
transports the bottle, aligns it with the shelf, places it, and releases it.}
{fig:sim_univtac_bottle_shelving}

\QualitativeTrajectory{0.96}
{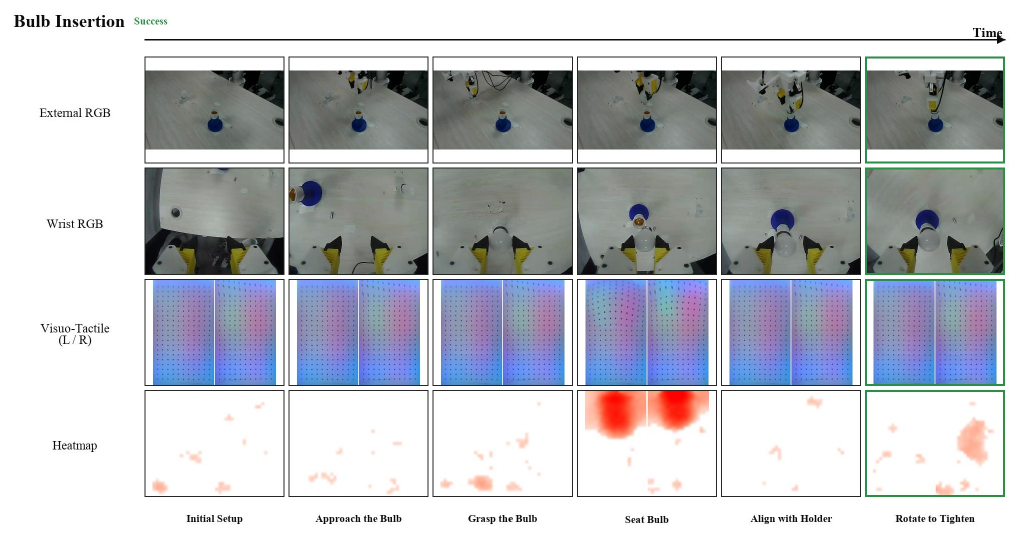}
{Successful real-robot bulb insertion. The final stages seat the bulb, align
it while maintaining the grasp, and rotate it to tighten.}
{fig:real_bulb_insertion}

\QualitativeTrajectory{0.96}
{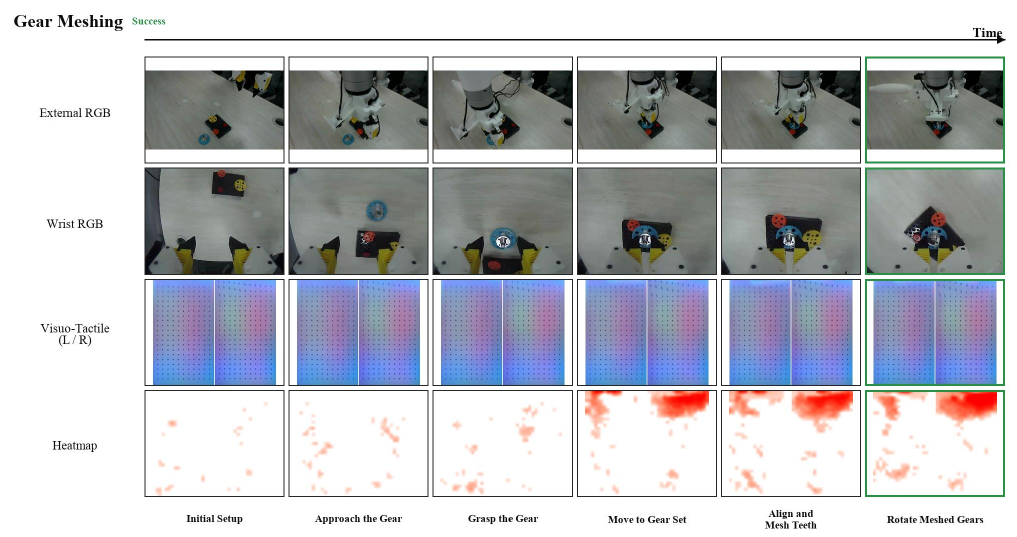}
{Successful real-robot gear meshing. After alignment and engagement, the
robot rotates the meshed gear to verify functional contact.}
{fig:real_gear_meshing}

\QualitativeTrajectory{0.96}
{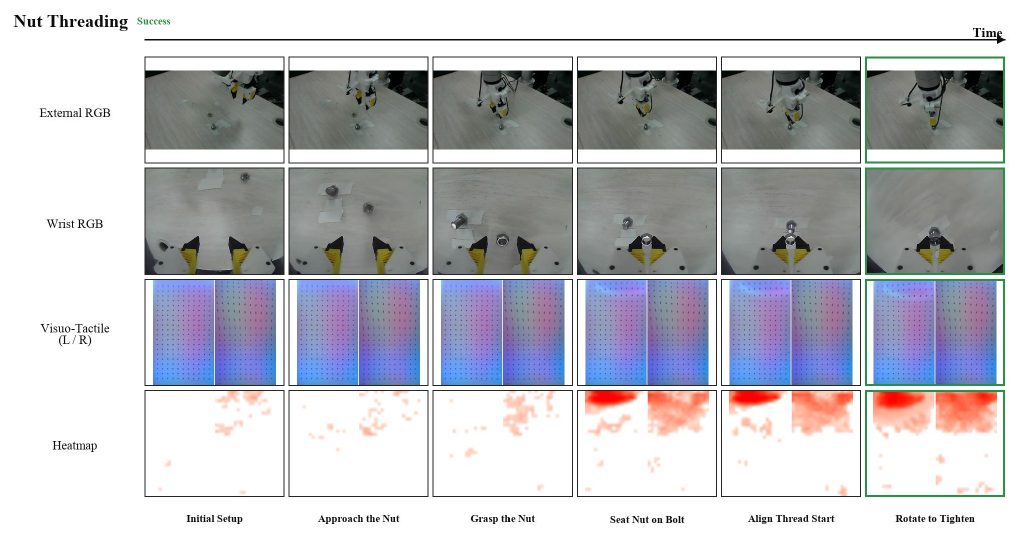}
{Successful real-robot nut threading, including approach, grasp, seating,
thread alignment, and tightening.}
{fig:real_nut_threading}

\QualitativeTrajectory{0.96}
{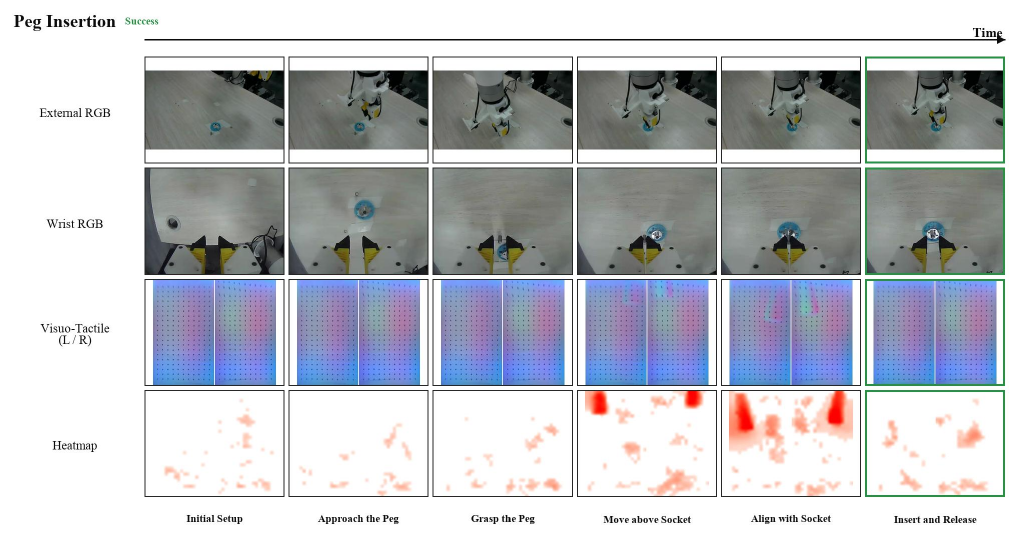}
{Successful real-robot peg insertion, from approach and grasp to socket
alignment and final insertion.}
{fig:real_peg_insertion}

\QualitativeTrajectory{0.96}
{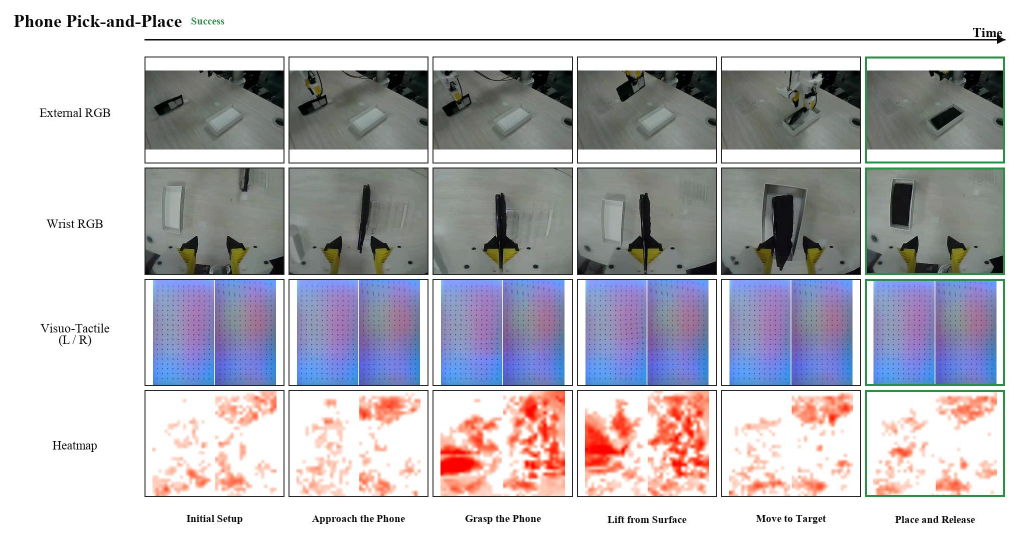}
{Successful real-robot phone pick-and-place execution, including grasp, lift,
transport, placement, and release.}
{fig:real_phone_pick_place}

\QualitativeTrajectory{0.96}
{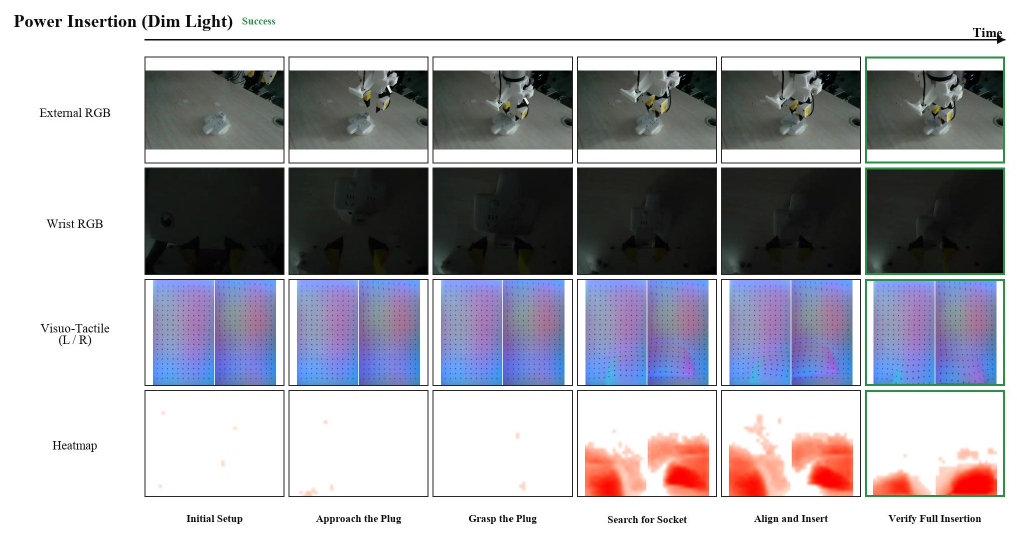}
{Successful real-robot power-plug insertion under dim lighting. The sequence
shows socket search, contact-guided alignment, insertion, and verification.}
{fig:real_power_insertion_dim}

\QualitativeTrajectory{0.96}
{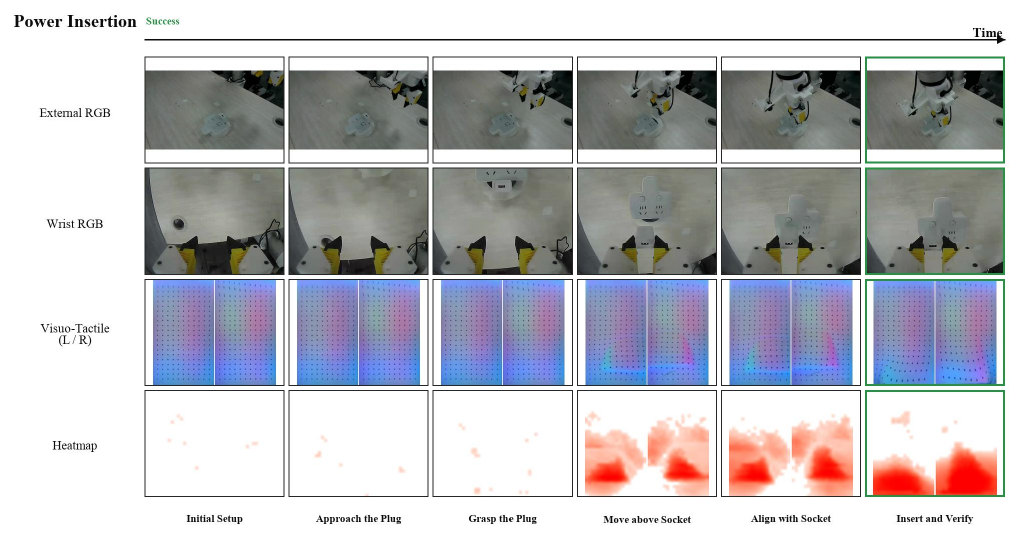}
{Successful real-robot power-plug insertion under standard lighting,
including socket alignment, insertion, and final verification.}
{fig:real_power_insertion}

\clearpage

\end{document}